\documentclass[sn-mathphys]{sn-jnl}

\jyear{2021}%

\theoremstyle{thmstyleone}%
\theoremstyle{thmstyletwo}%
\usepackage{soul, color, xcolor}

\theoremstyle{thmstylethree}%

\begin{document}

\title[MECPformer: Multi-estimations Complementary Patch for WSSS]{MECPformer: Multi-estimations Complementary Patch with CNN-Transformers for Weakly Supervised Semantic Segmentation}


\author[1]{\fnm{Chunmeng} \sur{Liu}}\email{lcm@tongji.edu.cn}

\author*[1]{\fnm{Guangyao} \sur{Li}}\email{97114@tongji.edu.cn}

\author[1]{\fnm{Yao} \sur{Shen}}\email{yaoshentj@tongji.edu.cn}

\author[1]{\fnm{Ruiqi} \sur{Wang}}\email{rqwang@tongji.edu.cn}

\affil*[1]{\orgdiv{ The College of Electronics and Information Engineering}, \orgname{Tongji University}, \orgaddress{\city{Shanghai}, \postcode{201804}, \country{China}}}


\abstract{
The initial seed based on the convolutional neural network (CNN) for weakly supervised semantic segmentation always highlights the most discriminative regions but fails to identify the global target information. 
Methods based on transformers have been proposed successively benefiting from the advantage of capturing long-range feature representations. However, we observe a flaw regardless of the gifts based on the transformer. Given a class, the initial seeds generated based on the transformer may invade regions belonging to other classes. 
Inspired by the mentioned issues, we devise a simple yet effective method with Multi-estimations Complementary Patch (MECP) strategy and Adaptive Conflict Module (ACM), dubbed \textbf{MECPformer}. Given an image, we manipulate it with the MECP strategy at different epochs, and the network mines and deeply fuses the semantic information at different levels. 
In addition, ACM adaptively removes conflicting pixels and exploits the network self-training capability to mine potential target information.
Without bells and whistles, our MECPformer has reached new state-of-the-art $72.0\%$ mIoU on the PASCAL VOC 2012 and $42.4\%$ on MS COCO 2014 dataset. The code is available at https://github.com/ChunmengLiu1/MECPformer.}




\maketitle

\section{Introduction}\label{sec1}

Semantic segmentation is a significant fundamental task for computer vision, which has rich application scenarios, such as autonomous driving~\cite{feng2020deep} and scene understanding~\cite{weng2021stage}.
Prior work has had great success in fully supervised semantic segmentation. 
However, pixel-level annotations are expensive and labor-intensive. Hence, weakly supervised semantic segmentation (WSSS) emerges with simpler annotations, e.g., bounding boxes~\cite{lee2021bbam}, scribbles~\cite{lin2016scribblesup}, points~\cite{bearman2016s} and image-level~\cite{ahn2018learning, lee2021railroad, wu2021embedded} labels. 
WSSS with the image-level label is the most accessible and challenging in the community. 
\emph{This paper concentrates on weakly supervised semantic segmentation (WSSS) with image-level labels.}

\begin{figure}[!t]
    \centering
     \includegraphics[scale=0.75]{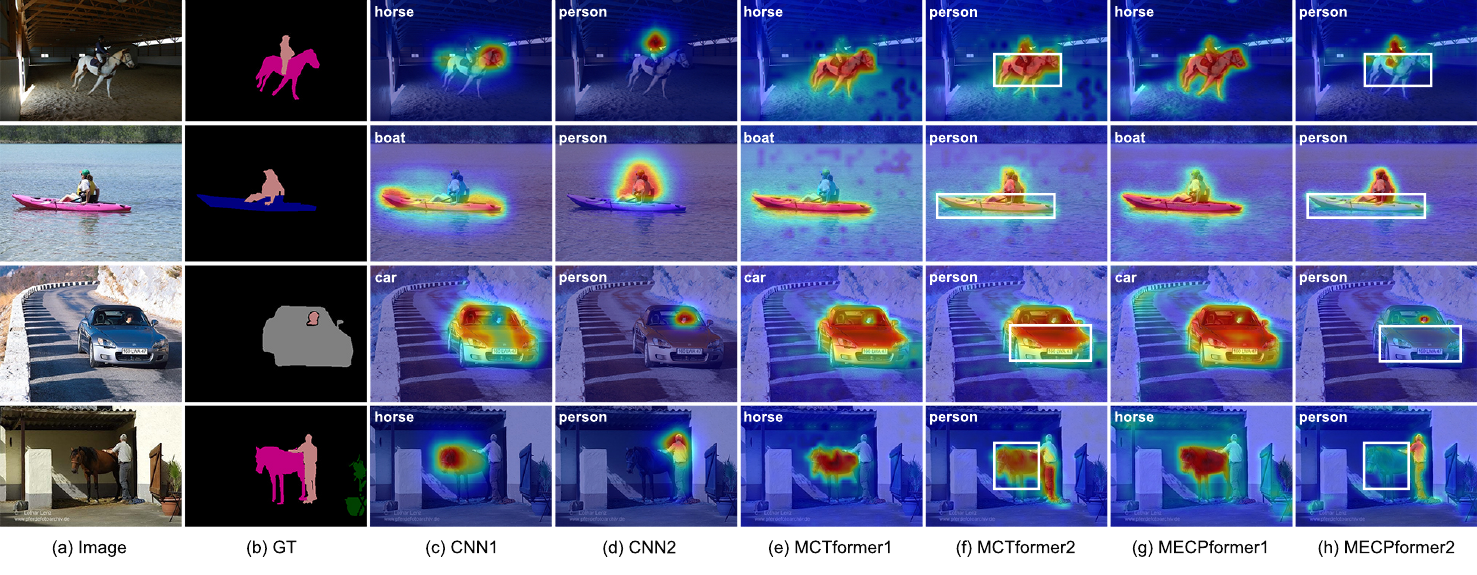}
    \caption{
    Comparison of the initial seeds of CNN, MCTformer \protect\cite{xu2022multi} and our MECPformer. 
    (c) and (d) are generated by ResNet38, while (e), (f) (g) and (h) are based on the Transformer. 
    Compared (c, d) and (e, f, g, h), we observe that the Transformer-based approach yields more integral activation regions compared to the CNN. The white bounding box indicates that the initial seed has invaded other categories. (g) and (h) illustrate that our MECPformer alleviates false-positive flaws. Zoom in for the best view}
    \label{Fig.heatmap}
\end{figure}

The currently standard WSSS methods have the following three steps:
1) Training a classification network with multi-label classification loss and generating the initial seeds. 2) Then post-processing methods refine the initial seeds, such as random erasing \cite{wei2017object}, and region expansion \cite{huang2018weakly}. 3) The segmentation network is self-trained with pseudo-labels.
The class activation map (CAM)~\cite{zhou2016learning} method is widely adopted to obtain the initial seeds.
However, we observed that CAM based on convolutional neural network (CNN) usually shows the most discriminative regions while missing the complete information about the object, as shown in Fig. \ref{Fig.heatmap} (c) and (d). One reason is that CNNs are subject to the limitation of capturing local information. An intuitive solution is to increase the perceptual field. However, it introduces gradient instability.
Current Transformer \cite{dosovitskiy2020image} has made new attempts and breakthroughs in a wide variety of computer vision tasks \cite{xie2021segformer, wang2021pyramid,peng2021conformer,li2022uniformer}. It has been proved to capture long-range dependent information due to the multi-head self-attention mechanism, which nature facilitates mining more complete object regions.
Benefiting from the advantages of Transformer, excellent works \cite{li2022transcam, xu2022multi} are proposed.

However, we observe the challenge of \emph{false positive pixels} of Fig. \ref{Fig.heatmap} (e) and (f), i.e., mislabeling information that is not relevant to the target class. Specifically, when A is the target class, CAM violates the irrelevant class B. 
As shown in Fig. \ref{Fig.heatmap} (f), the white boxes mark the unrelated classes. The response of irrelevant class B introduces noise, which is detrimental to the pseudo label.

Inspired by the above observations, we propose an innovative method, termed MECPformer, focusing on the initial seed generation and refinement pseudo-labeling steps.
%
Our MECPformer is a dual branch network, in which each consists of a Conformer \cite{peng2021conformer}, and weights shared. 
Specifically, we introduce a Multi-estimations Complementary Patch (MECP) strategy for the input images in different training epochs.
When discriminative regions of the input image are occluded, the network tries to find other response regions associated with the category label. Multiple estimations enable the network to focus on different levels of semantic information, further enhancing the learning ability of the input images.  
The MECP strategy enhances the network's ability to recognize target objects and obtain complete and accurate semantic objects. It is worth noting that the MECP strategy is only applied in the training phase.
To further overcome the disadvantage of false positives, we present a new strategy for post-processing named Adaptation Conflict Module (ACM).
ACM detects conflicting pixels in all category's initial seeds and marks conflicting pixels as ignored pixels 255 employing adaptive processing. In addition, the self-learning capability of the segmentation network mines potential information at conflicting pixel locations to yield higher-quality prediction results.
Extensive experiments on the PASCAL VOC 2012 \cite{everingham2015pascal} dataset demonstrate the superiority of our MECPformer. When it comes to the VGG16 \cite{simonyan2014very} backbone in the self-training stage, the mIoU of MECPformer achieves $67.7\%$. When performing experiments on another ResNet101 \cite{he2016deep} backbone, we show a test set result of $72.0\%$. It reaches $42.4\%$ when benchmarked on the larger MS COCO 2014 dataset.

In summary, our main contribution is three-fold:
\begin{itemize}
    \item We design the Multi-estimations Complementary Patch (MECP) strategy in MECPformer to dig and merge different levels of semantic information in different training epochs to mitigate the false positive defects.
    
    \item Adaptation Conflict Module (ACM) is proposed as a novel post-processing method to further alleviate the problem of conflicting initial seeds of different classes brought about by false positive drawbacks.
    
    \item Experiments on PASCAL VOC 2012 and MS COCO 2014 dataset demonstrate that our method outperforms previous methods and achieves superior performance.

\end{itemize}

\section{Related Work}
We briefly overview initial seed generation, initial seed refinement, and False Positive Flaws in WSSS.
\subsection{Initial Seeds Generation}
Most existing WSSS methods generate initial seeds from the CAM extracted from the CNN-based classification network, followed by refinement of the initial seed to train the segmentation network. 
The initial CAM invariably has coarse boundary profiles and is insufficiently complete. A wide range of scholars has identified that only the most discriminative regions are activated and contributed to many marvelous endeavors to remedy this limitation. 
The erasure method removes the most discriminative regions from the initial CAM, forcing the network to concentrate on the remaining target regions \cite{wei2017object, hou2018self}. 
Researchers fuse the diverse CAM yielded by the dilated convolution of various dilated rates \cite{wei2018revisiting,lee2019ficklenet}.
OAA \cite{jiang2019integral} integrates CAM of different training processes to achieve a complete activation region. 
SC-CAM \cite{chang2020weakly} is equipped with a self-supervised clustering algorithm to mine the implicit subcategory information.
SIPE \cite{chen2022self} adopts a self-supervised manner for image-specific prototype exploration.
MCIS \cite{sun2020mining} considers the semantics between images internally for cross-image information mining.
Hide-and-Seek \cite{kumar2017hide} hides image patches randomly against the input image, enabling the network to locate more relevant regions. Inspired by Hide-and-Seek, CPN \cite{zhang2021complementary} presents the complementary patch network dedicated to excavating more foreground regions.
The mentioned works have significantly contributed to WSSS by depending on the advantage of CNN's capability to draw local features. However, they suffer from the limitation of insufficient ability to acquire long-distance dependent information.

ViT \cite{dosovitskiy2020image} first introduces the Transformer to computer vision (CV). Following that, Transformer excelled in many fundamental CV works, such as semantic segmentation \cite{xie2021segformer}, depth estimation \cite{ranftl2021vision}, and video understanding \cite{arnab2021vivit}.
Since Transformer holds the advantage of catching global information, Transformer has also been taken into WSSS. The AFA \cite{ru2022learning} adopts one-stage training to acquire semantic affinity from self-attention mechanisms. The preceding work opens up a new era for WSSS. MCTformer \cite{xu2022multi} proposes a multi-class token Transformer, demonstrating the importance of tokens for Transformers.

The fusion of local knowledge with global characteristics has also drawn increasing attention \cite{peng2021conformer, li2022uniformer}. Conformer \cite{peng2021conformer} elaborates the FCU module to communicate the information from both CNN and Transformer branches. UniFormer \cite{li2022uniformer} devises a versatile module to integrate the feature information from CNN and Transformer serially. TransCAM \cite{li2022transcam} first presented Conformer as the backbone of WSSS. 

However, we observed that works based on Transformers suffer from false positive flaws. We mitigate the issue significantly by adopting the MECP.

\subsection{Initial Seeds Refinement}
Certain efforts concentrate on the refinement of the initial pseudo tags.
Contrast loss \cite{ke2021universal}, SEC loss \cite{kolesnikov2016seed} and CRF loss \cite{zhang2020reliability} strengthen the constraints on the segmentation phase. DSRG\cite{huang2018weakly} utilizes deeper feature information in the segmentation phase with seed region growth. AffinityNet \cite{ahn2018learning} performs semantic propagation across the predicted semantic affinities using random walk. Some work \cite{kim2021discriminative, lee2021railroad, wu2021embedded, yao2021non} obtain distinct pseudo-label boundaries by taking the saliency map as background information. 
Enlightened by the above work, we propose a simple and effective pseudo labels post-processing method called Adaptation Conflict Module (ACM). It adaptively mitigates the conflicting pixels between classes and exploits the self-learning capability of the segmentation network for potential information mining. It also contributes to false positive defects.

\subsection{False Positive Flaws}
False positive defects have become a significant concern for researchers working on WSSS. Many studies have been conducted based on CNNs. Augmenting the capability to extract pertinent target features \cite{wei2018revisiting} and optimize the loss function \cite{wang2019self,wang2020self} during the classification phase constitutes a means of diminishing false-positive defects.
CIAN \cite{fan2020cian} introduces a cross-image affinity module that effectively suppresses false positive predictions. ReCAM \cite{chen2022class} replaces the Binary Cross Entropy (BCE) loss function with Softmax Cross Entropy (SCE) to generate high-quality pseudo labels with fewer false positives. ESOL \cite{li2022expansion} proposes a shrinkage sampler that leverages an additional deformable convolutional layer to exclude false positive regions with loss minimization optimally.
In addition, post-processing techniques have proven to be effective in mitigating false positives \cite{roy2017combining}.
DCSP \cite{chaudhry2017discovering} utilizes the saliency map to remove false positives and improve pseudo-labeling accuracy. AffinityNet \cite{ahn2018learning} employs random walk to effectively handle false positives in CAM. InferCam \cite{sun2022inferring} introduces a refined inference-only iterative module that can further reduce false-positive defects. URN \cite{li2022uncertainty} performs uncertainty estimation of noise suppression by response scaling to reduce false positive pixels.

The Transformer-based approach investigates false positives by integrating attention mechanisms \cite{yu2022ex}. AD-CAM \cite{huang2022attention} uses a new probabilistic diffusion coupling method that integrates ViT attention and CAM activation, alleviating false positives caused by the Conformer.
SemFormer \cite{chen2022semformer} evaluates class-relative attention (CRA) and significantly improved segmentation performance.
MCTformer \cite{xu2022multi} reduces the number of false positives by reducing the number of layers of attentional fusion.
In this work, we address false positives other than the attention mechanism. Specifically, our approach leverages the MECP strategy during the classification stage to augment the network's image feature extraction capabilities. We also introduce the ACM post-processing algorithm, which identifies conflicting pixels of different categories within the initial seed and assigns them a value of 255 to further refine false positives. By adopting a dual approach, we effectively mitigate the issue of false positives and achieve superior performance compared to existing techniques.

\section{Methodology}
\begin{figure*}[!t]
    \centering
     \includegraphics[scale=0.32]{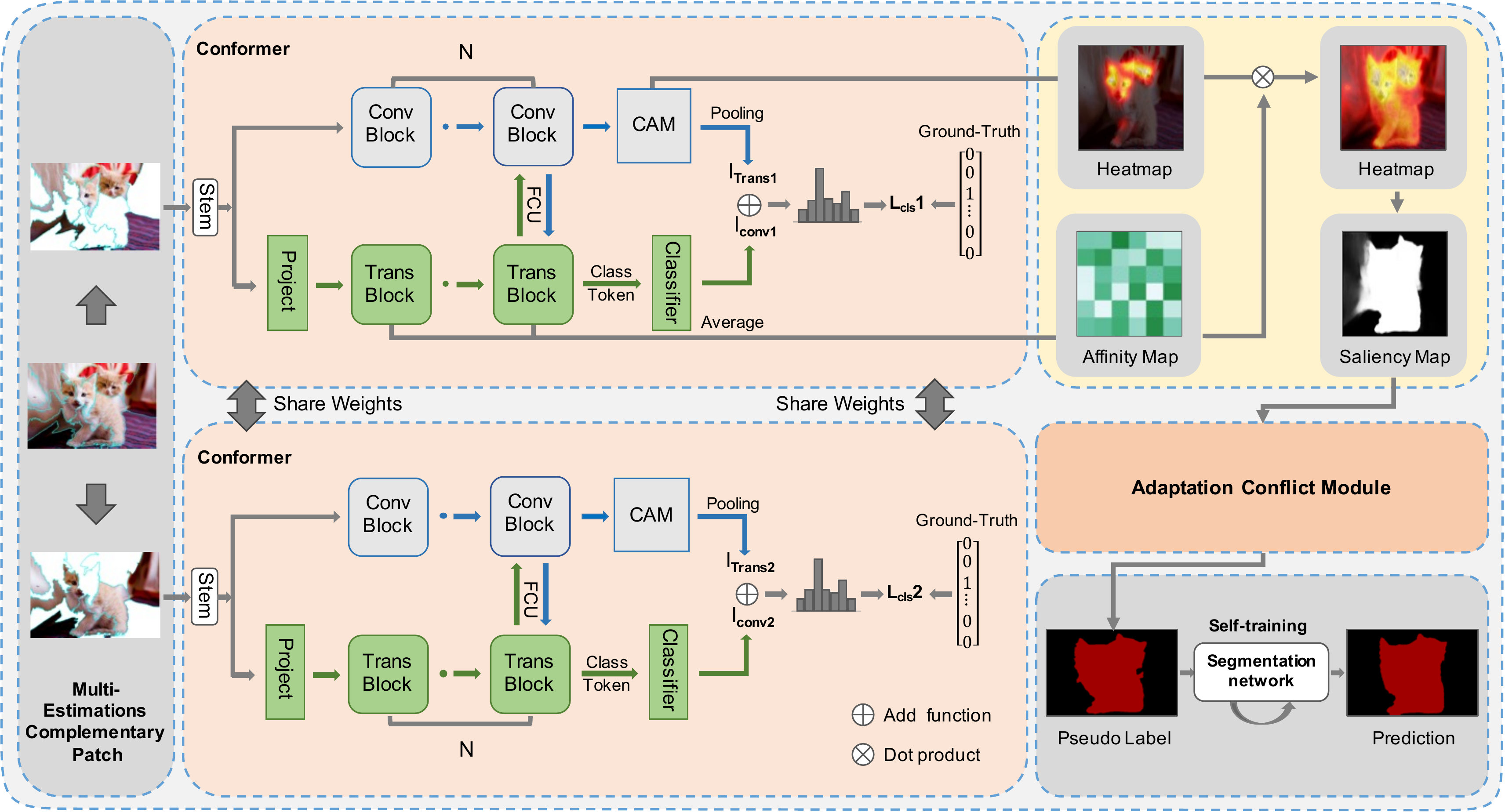}
    \caption{Overview pipeline of MECPformer.
    First, the Multi-estimations Complementary Patch (MECP) strategy produces two complementary patch images to feed into the classification network.
    Second, we train the classification network with multi-label classification loss to obtain initial seeds and affinity map, and We integrate them to get refined initial seeds.
    Third, the Adaptation Conflict Module~(ACM) adaptively filters conflicting pixels to avoid false pseudo labels and filter out excess background information with saliency maps.
    Finally, the generated credible pseudo labels are utilized for training the segmentation network, and the network's self-learning capability is exploited to create more delicate prediction results}
    \label{Fig.framework}
\end{figure*}

The overview framework of MECPformer represents in Fig.~\ref{Fig.framework}. MECPformer is two-branched, and each branch, in turn, consists of Conformer \cite{peng2021conformer}.
First, the MECP strategy manipulates the image into two complementary patches, which serve as input in the training phase. We adopt multiple estimations complementary patches at different training epochs.
Second, we train the classification network with multi-label classification loss to generate initial seeds and affinity maps. 
Furthermore, the adaptation conflict module adaptively ignores conflicting pixels and presents dense and trustworthy pseudo labels.
Finally, the self-learning capability of the segmentation network is fully exploited to mine the potential feature information of the target and yield more refined and integrated prediction results.
\subsection{Multi-estimations Complementary Patch Strategy}
\begin{figure}[!t]
    \centering
     \includegraphics[scale=0.4]{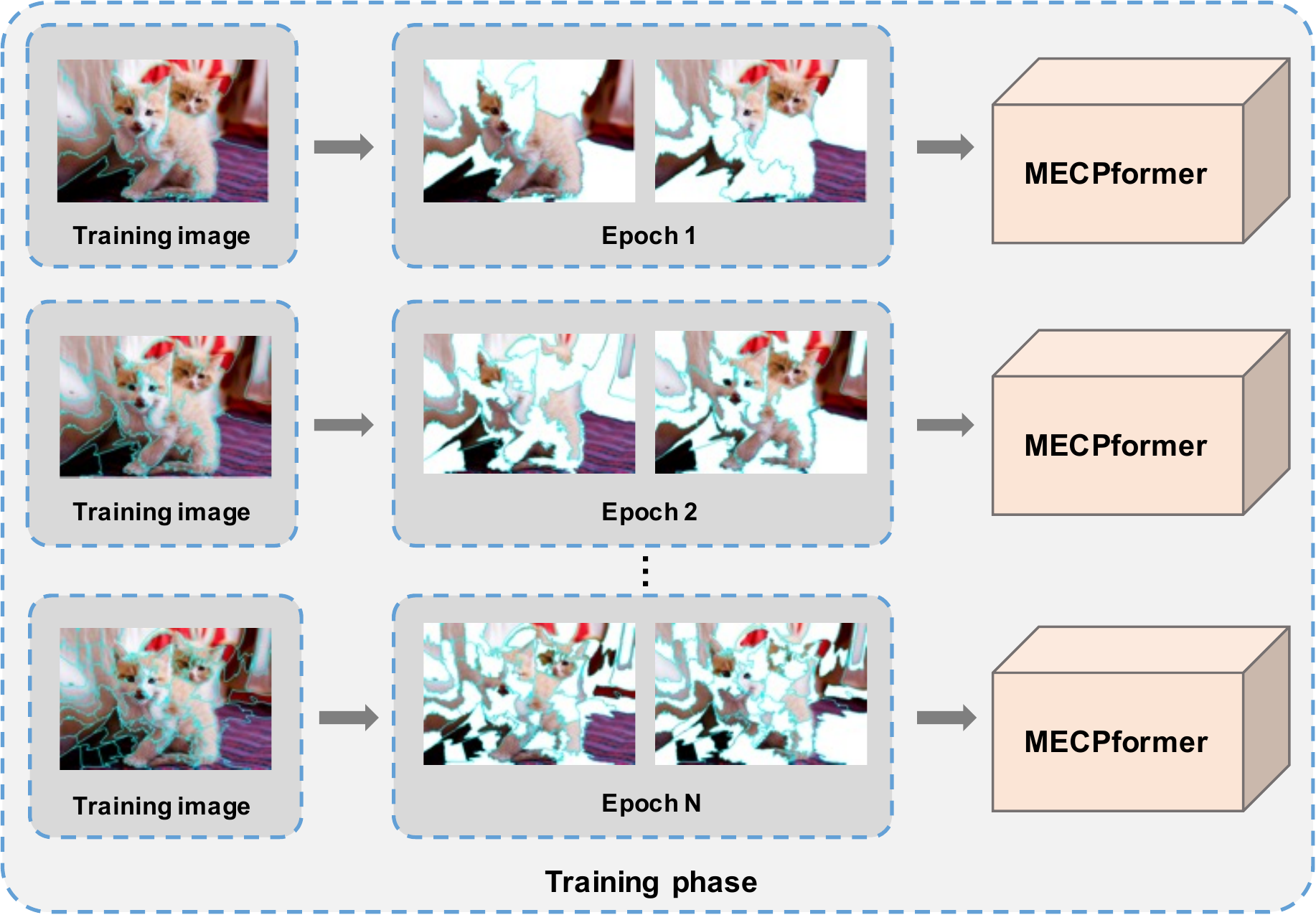}
    \caption{Overview of Multi-estimations Complementary Patch strategy (MECP). Multiple estimations are taken for different epochs to obtain irregular image patches of various sizes. After receiving pairs of complementary patches, they are fed to MECPformer (ours) as input for training}
    \label{Fig.mecp}
\end{figure}

The MECP strategy yields pairs of complementary training images, as depicted in Fig.~\ref{Fig.mecp}.

Let's represent an initial input image as $I\in \mathbb{R}^{C\times H\times W}$. Complementary patches are denoted as $I_{cp}\in \mathbb{R}^{C\times H\times W}$, $I_{\bar{cp}}\in \mathbb{R}^{C\times H\times W}$. Notably, 
\begin{equation}
I_{cp} + I_{\bar{cp}} = I.
\end{equation}

Benefiting from the fast computational speed and efficient memory of the simple linear iterative clustering (SLIC) \cite{achanta2012slic}, we take it as the basis for MECP. SLIC adapts the clustering method, and $k$ dictates the number of clustering centers. By default, $k$ is the only hyper-parameter of the SLIC. The clustering process begins with initializing the clustering centers by sampling from a regular grid spaced by $g$ pixels. To produce superpixel patches of approximately the same size, the grid interval is $g=\sqrt{\frac{G}{k}}$, where $G$ is the total number of pixels. For more details, please see SLIC \cite{achanta2012slic}.

It is worth noting that $k$ irregular semantic patches are obtained for each image after SLIC. Each patch carries the same probability $p_r$ being hidden and $p_r$ is usually considered to be $0.5$. After that, we found that the super-pixel patch is rich in semantic information, which inspired us to fuse images with different segmentation numbers $k$ during the training process. As illustrated in Fig.~\ref{Fig.mecp}, the MECP strategy is employed in the training phase and is excluded during testing. Different epochs are set with different estimations, i.e., $k$ takes different values. Various $k$ values divide the images into patches of different sizes. The fusion of features with different granularity makes the network picks up the target features from different perspectives. After obtaining the complementary patch pairs, they are fed into our classification branch networks respectively.

\subsection{Initial Seed Generation}
\subsubsection{Network Structure}
Each branch comprising the MECPformer is the Conformer \cite{peng2021conformer}, as shown in Fig. \ref{Fig.framework}. The Conformer is a parallel two-branch network consisting of a CNN and a Transformer branch, respectively. 

The CNN branch is a pyramid structure, and while the network depth increases, the scale of the feature map decreases. 
There are the same numbers $N$ of convolutional blocks as Transformer blocks. 
The CNN branch adopts convolution operations instead of linear layers for classification. CNN branch benefits from convolutional operations that successively supply the network with local detail information.

The Transformer branch is designed to follow the ViT \cite{dosovitskiy2020image}. 
The input image $I$ is sliced into $\sqrt{s}\times \sqrt{s}$ patches, and the embedding dimension is denoted as $D$.
Each transformer block contains the multi-head self-attention mechanism module and MLP. Class token learned information is fed to the classification head for categorization. The Transformer branch contributes significantly to the network acquisition of global context representation information. 

The Conformer proposes the feature coupling unit (FCU) module to communicate to take full advantage of local features and global dependencies. 

\subsubsection{Class Activation Map Generation}
The classification activation map (CAM) is widely adopted as an initial seed generation method. Considering the simplicity and effectiveness of the CAM method, we choose CAM to generate the initial seed in the CNN branch.

By convention, the input image $I \in \mathbb{R}^{C\times H\times W}$ is fed into the CNN branch to yield the feature map $f \in \mathbb{R}^{f_c\times f_h\times f_w}$, where $f_h\times f_w$ denote spatial dimension, $f_c$ is the channel number.
Further, a $1\times 1$ convolutional layer with a learnable matrix $w\in\mathbb{R}^{f_c}$ and global average pooling operates on $f$ to derive the prediction logits $L_{conv}$. The initial CAM $M_b\in\mathbb{R}^{f_h \times f_w}$ is then obtained by multiplying the transpose of the learnable matrix $w$ with the feature map $f$. $b$ is a specific class. The equation is expressed as follows:
\begin{equation}
    M_b=w^T f,
\end{equation}
where $f$ indicates the feature map.

Due to the limitation of CNN, here CAM shows the most discriminative regions, which lacks integrity and makes it challenging to obtain optimal pseudo labels. Inspired by TransCAM~\cite{li2022transcam}, global contextual information in the multi-head self-attention module contributes a lot to refining CAM.

\subsubsection{Affinity-Based CAM Refinement}
Researchers note the consistency between the multi-head self-attention (MHSA) in Transformers and the semantic level of affinity \cite{ru2022learning,li2022transcam}. Inspired by them, we extract semantic affinity maps from the MHSA module.

As shown in Fig.~\ref{Fig.framework}, there are $N$ Transformer blocks in the Transformer branch, and MHSA is in each Transformer block $n$.
Specifically, for the $n$ block, $Q^n \in \mathbb{R}^{(1+s)\times D}$, $K^n \in \mathbb{R}^{(1+s)\times D}$ and $V^n \in \mathbb{R}^{(1+s)\times D}$ mean query, key and value of representations projected from $I^n$. $I^n$ denotes the input of the MHSA. Based on $Q^n$, $K^n$, and $V^n$, the affinity map $A_n$ and output $O_n$ are defined as follows:
\begin{equation}
  A^{n} = \text{softmax}(\frac{{Q^{n}\cdot K^{n}}^T}{\sqrt{D}}),
\end{equation}
\begin{equation}
  O^{n} = A^{n}\cdot V^n,
\end{equation}
where $A^{n}\in \mathbb{R}^{(s+1)\times (s+1)}$, here $s$ is the number of image patch tokens and $1$ is the class token.
We calculate the average affinity map across all Transformer blocks.
\begin{equation}
  \bar{A} = \frac{1}{N}\sum_n(A^{n}).
\end{equation}
The final affinity map $\bar{A}\in \mathbb{R}^{(s+1)\times (s+1)}$ enriches the long-range feature information. To make full use of the local information with global semantic affinity, we perform matrix multiplication of CAM $M^r_b$ and affinity map $\bar{A}^{\star}$ to obtain $\widehat{M_b}\in \mathbb{R}^{s\times 1}$ with the following equation: 
\begin{equation}
    \widehat{M_b}=M^r_b\cdot {\bar{A}^{\star}}.
\end{equation}
For $M_b\in\mathbb{R}^{f_h \times f_W}$, $f_h$ and $f_w$ are each equals to $\sqrt{s}$. $M^r_b \in \mathbb{R}^{s\times 1}$ is derived from the $M_b$ reshaping. $\bar{A}\in \mathbb{R}^{(s+1)\times (s+1)}$ is given by eliminating the information associated with the class token to get $\bar{A}^{\star}\in \mathbb{R}^{(s)\times (s)}$. The $\widehat{M_b}$ is then reshaped into $\widehat{M_b}\in\mathbb{R}^{\sqrt{s} \times \sqrt{s}}$. $\widehat{M_b}$ sufficiently consolidates local features and global contextual information to derive a more integral initial seed.

\subsection{Adaptation Conflict Module}

We put forward a simple and effective post-processing module Adaptation Conflict Module (ACM). 

The detail of ACM is illustrated as follows:
\begin{equation}
\label{equ.finlabel}
  P_{(x, y)}=\left \{\begin{array}{ll}
  255, & {E^{fir}_{(x, y)}}~>~1\\255, & {E^{sec}_{(x, y)}}~>~0\\ P_{(x, y)}, & \text{otherwise.} \end{array}\right.
\end{equation}
$P_{(x, y)}\in \mathbb{R}^{H\times W}$ is the initial pseudo label, which is obtained by comparing the initial seeds of the whole classes $B$ with the $argmax$ function. 
Equation \ref{equ.finlabel} indicates that in pixel $(x, y)$ position, if $E^{fir}_{(x, y)}$ is greater than 1 or $E^{sec}_{(x, y)}$ is larger than 0, the pixel at $(x, y)$ marks as 255, otherwise the pixel remains constant.

The $E^{fir}_{(x, y)}$ is defined follows:
\begin{equation}
  \widehat{P}_\text{max} = \text{max}(\widehat{P}),
\end{equation}
\begin{equation}
  P_\text{exp} = \text{expand}(\widehat{P}_\text{max}, \widehat{P}),
\end{equation}
\begin{equation}
  C_m =  P_\text{exp}\odot C_r,
\end{equation}
\begin{equation}
  E^{fir} = \sum(\widehat{P} > C_m, axis=0).
\end{equation}
$\widehat{P}\in\mathbb{R}^{B\times H \times W}$ are the initial seeds without $argmax$ function. 
 First, we find the $\widehat{P}$ maximum $\widehat{P}_\text{max}\in \mathbb{R}^{H\times W}$ over sum classes $B$ channels. Then we expand $\widehat{P}_\text{max}$ to the same dimension as $\widehat{P}$ to obtain  $P_\text{exp}\in\mathbb{R}^{B\times H \times W}$. After that, we multiply $P_\text{exp}$ by the conflict rate $C_r$ to get the conflict matrix $C_m\in\mathbb{R}^{B\times H \times W}$. Eventually, we compare $\widehat{P}$ and $C_m$ and sum in channels $B$. Equation \ref{equ.finlabel} calculates the conflicting pixels between the initial seeds of different categories. If more than two categories of initial seeds exist simultaneously at the same location, the location is considered to be conflicting pixels and marked as $255$.  

$E^{sec}_{(x, y)}$ is expressed in the following form:
\begin{equation}
\label{equ.E2}
  E^{sec} = (S >= T_{bg})\odot(\bar{P}_{(x,y)} == 0),
\end{equation}
where $S$ means saliency map. $T_{bg}$ is a background threshold value in the range of 0 to 255.
$\bar{P}_{(x,y)}$ is the pseudo label $P_{(x,y)}$ processed by $E^{fir}$.
Equation \ref{equ.E2} demonstrates another form of conflict at a pixel $(x,y)$.
When the saliency map considers a valid class (except for the background class), yet the pseudo label $\bar{P}_{(x,y)}$ treats it as a background class, we label the pixel at $(x, y)$ as 255.

Following the above operation, we remove the uncertain pixels in the initial pseudo labels and yield more trustworthy pseudo labels.

\subsection{Self-training with Pseudo Label}
Subsequent to the adaptation conflict module, we achieve intensive and trusted pseudo labels. 
Then, we exploit the potential target information with the self-learning capability of the segmentation network. 
The experimental results demonstrate that we attain more refined and complete prediction results. 

\section{Experiments}

\subsection{Experimental Settings}

\subsubsection{Dataset and Evaluation Metrics}
We evaluate our segmentation results on the PASCAL VOC 2012 \cite{everingham2015pascal} and MS COCO 2014 \cite{lin2014microsoft}. \textbf{PASCAL VOC} has 20 foreground semantic object classes and one background class. It is separated into three subsets, 1464 images for training, 1449 images for validation, and 1456 for testing. Following the existing work experience, we adopt an additional augmented dataset from SBD \cite{hariharan2011semantic} with 10582 images for training. PASCAL VOC test results are available on its official evaluation website. \textbf{MS COCO} contains 80 foreground classes and one background class, with 80k and 40k images in the training and validation sets, respectively.
We choose mean Intersection-over-Union~(mIoU) as our evaluation metric.

\subsubsection{Implementation Details}
The model is trained on 1 NVIDIA GeForce RTX 3090 GPU with 24 GB memory.
Our method pays attention to the classification phase based on the Conformer-S \cite{peng2021conformer} backbone, which we employ pre-trained weights on the ImageNet \cite{russakovsky2015imagenet} and fine-tune them on the PASCAL VOC 2012 with image-level labels. For training, the input image is scaled randomly from $320$ to $640$ and crop them into the size of $512\times512$.
We use the AdamW optimizer with the learning rate 5e-5. Batch size is set to 4 while Weight decay equals 5e-4. When inferring, aggregating the input image results for various sizes ($256\times256$, $512\times512$, $768\times768$) contributes significantly to the final result. It is worth noting that the MECP strategy is not applied in the inference phase.

For the segmentation step, followed the common practice \cite{yao2021non,kim2021discriminative,jiang2019integral}, we adopt the DeepLab-v2 backbone and present our evaluation results on ResNet101 and VGG16. We deployed an SGD solver with weight decay 5e-4. We pre-trained on MS-COCO, and the final result is after CRF \cite{chen2017deeplab}.

\subsubsection{Model Complexity}
We present a comparative analysis between MECPformer and TransCAM~\cite{li2022transcam} against ResNet38~\cite{wu2019wider}, a widely-used CNN model for generating initial seeds \cite{ahn2018learning, wang2020self, zhang2021complementary}, based on image inference metrics including image size, number of parameters, and multiply-add calculations (MACs). Table \ref{Tab.complex} displays the results of our evaluation. Notably, our findings demonstrate that the proposed Conformer-S-based approach exhibits significantly lower complexity in comparison to ResNet38.
\begin{table}[ht]
\begin{center}
\begin{minipage}{260pt}
\caption{Complexity of models generation initial seeds. The proposed MECPformer is based on Conformer-S~\cite{peng2021conformer}. MS means multi-scale inference ($256\times256$, $512\times512$, $768\times768$).}
\label{Tab.complex}
\begin{tabular*}{\textwidth}{@{\extracolsep{\fill}}lccc@{\extracolsep{\fill}}}
\toprule
 \rule{0pt}{9pt} \textbf{Model}&\textbf{Image size}&\textbf{Params(M)}&\textbf{MACs(G)}\\
\midrule
 \rule{0pt}{9pt}  \text{ResNet38~\cite{wu2019wider}} & \text{$224\times224$} & \text{104.3} &  \text{99.8}\\
 \rule{0pt}{9pt}  \text{TransCAM~\cite{li2022transcam}} & \text{MS} & \text{36.5} &  \text{13.9}\\
 \rule{0pt}{9pt} \text{MECPformer~(ours)}  & \text{MS} & \text{36.5} & \text{14.3}\\
 \bottomrule
\end{tabular*}
\end{minipage}
\end{center}
\end{table}

\subsection{Comparisons to the State-of-the-arts}
\subsubsection{PASCAL VOC}

\begin{table}[htbp]
\begin{center}
\begin{minipage}{260pt}
\caption{ MIoU evaluation results based on the initial seeds on the PASCAL VOC 2012 train set }
\label{Tab.ini_seed}
\begin{tabular*}{\textwidth}
{@{\extracolsep{\fill}}lccc@{\extracolsep{\fill}}}
\toprule
 \rule{0pt}{9pt}   \textbf{Method}   & \textbf{Pub.} & \textbf{Backbone} & \textbf{Train} \\
\midrule
 \rule{0pt}{9pt}   \text{SC-CAM} \cite{chang2020weakly}  & \text{CVPR20} & \text{ ResNet101}  & \text{50.9}\\
 \rule{0pt}{9pt}   \text{SEAM} \cite{wang2020self} & \text{CVPR20}   & \text{ResNet38}  & \text{55.4} \\
 \rule{0pt}{9pt}   \text{AdvCAM} \cite{lee2021anti} & \text{CVPR21} & \text{ ResNet101} & \text{55.6} \\
 \rule{0pt}{9pt}   \text{CPN} \cite{zhang2021complementary} & \text{ICCV21}  & \text{ResNet38} & \text{57.4}  \\
 \rule{0pt}{9pt}   \text{AMR} \cite{qin2022activation} & \text{AAAI22}  & \text{ResNet50} & \text{56.8}  \\
 \rule{0pt}{9pt}   \text{SIPE} \cite{chen2022self} & \text{CVPR22}  & \text{ ResNet101} & \text{58.6} \\
 \rule{0pt}{9pt}   \text{MCTformer} \cite{xu2022multi} & \text{CVPR22}  & \text{ DeiT-S} & \text{61.7} \\
   \rule{0pt}{9pt}   SemFormer \cite{chen2022semformer}  & Arxiv22 & Deit-S & 63.7\\
 \rule{0pt}{9pt}   AD-CAM \cite{huang2022attention}  & Arxiv22 & Conformer-S & 63.5\\
 \rule{0pt}{9pt}   \text{TransCAM} \cite{li2022transcam}  & \text{Arxiv22} & \text{Conformer-S} & \text{64.9}\\
\midrule
\rule{0pt}{9pt}   \textbf{MECPformer (Ours)}  & \text{-} & \text{Conformer-S} & \textbf{66.6}  \\
\bottomrule
\end{tabular*}
\end{minipage}
\end{center}
\end{table}
\begin{figure}[htbp]
    \centering
    \scalebox{1}{\includegraphics[scale=1]{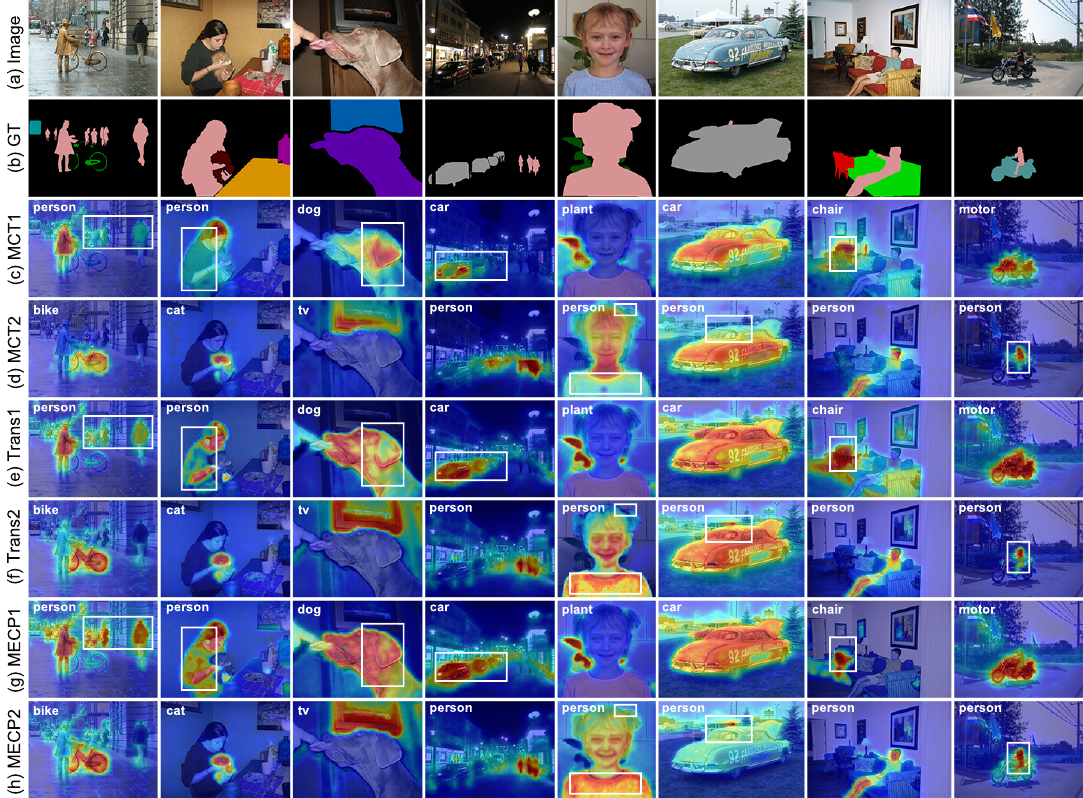}}
    \caption{Qualitative comparison of initial seeds.``MCT'' means MCTformer \protect\cite{xu2022multi} method and ``Trans'' denotes TransCAM \protect\cite{li2022transcam}. ``MECP'' is our MECPformer. Our MECPformer has a stronger grasp of the target features and produces a more dense and plausible initial seed
    }
    \label{Fig.repre_heat}
\end{figure}

\textbf{Improvement on Initial Seeds.} We first evaluate the quality of the initial pseudo labels. Following the common practice \cite{chen2022self, qin2022activation,zhang2021complementary}, we mainly present the results of the PASCAL VOC 2012 training set. As shown in Table~\ref{Tab.ini_seed}, we have a significant performance improvement over the current prevalent methods. Compared to SIPE \cite{chen2022self}, we have an $8.0\%$ boost.
When compared to MCTformer \cite{xu2022multi}, we outperform their results by $4.9\%$.
When comparing with TransCAM \cite{li2022transcam}, which also employs the conformer structure, our results outperform theirs by $1.7\%$.
According to the results of mIoU, our MECPformer yields higher-quality initial seeds and alleviates the false positive flaws. We have performed a qualitative demonstration of the initial seeds in Fig. \ref{Fig.repre_heat}. As can be seen, our MECPformer is better able to identify class-specific features and produce more dense and plausible initial seeds.

\textbf{Improvement on Segmentation Results.} To further evaluate the performance of our method, we show a segmentation performance comparison with the latest excellent methods. 
Table~\ref{Tab.sota_vgg_imn} and Table~\ref{Tab.sota_resnet} present the results for VGG16 and ResNet101 backbone on the PASCAL VOC 2012 dataset, respectively. 
In Table~\ref{Tab.sota_vgg_imn} of the VGG16 backbone, our segmentation results reach $67.7\%$ and $67.4\%$ on the validation set and test set, respectively. Specifically, 
our MECPformer results are $4.1\%$ ahead of DRS \cite{kim2021discriminative} and $2.2\%$ above NSROM~\cite{yao2021non}. Furthermore, our results exceed EPS~\cite{lee2021railroad} $0.7\%$ while outperforming $I^{2}CRC$ \cite{chen2022saliency} $3.4\%$. 
In addition, compared with those methods without saliency maps,
the proposed MECPformer improves upon the best performance by over $5.6\%$.
\begin{table}[ht]
\begin{center}
\begin{minipage}{260pt}
\caption{Quantitative comparisons to previous state-of-the-art approaches with VGG16 backbone in segmentation network in the self-training stage. The result is with CRF}
\label{Tab.sota_vgg_imn}
\begin{tabular*}{\textwidth}{@{\extracolsep{\fill}}lcccc@{\extracolsep{\fill}}}
\toprule
 \rule{0pt}{9pt}\textbf{Methods} &\textbf{Backbone}  & \textbf{Sup.} & \textbf{Val}  & \textbf{Test}  \\ \midrule
 \rule{0pt}{9pt}AffinityNet~(CVPR18)~\cite{ahn2018learning} & VGG16 & I  & 58.4   & 60.5     \\
 \rule{0pt}{9pt}BES~(ECCV20)~\cite{chen2020weakly} & VGG16  & I  & 60.1   & 61.1        \\
 \rule{0pt}{9pt}ECS-Net~(ICCV21)~\cite{sun2021ecs} & VGG16  & I  & 62.1   & 63.4        \\
 \hline
 \rule{0pt}{9pt}GAIN~(CVPR18)~\cite{li2018tell} & VGG16 & I+S   & 55.3   & 56.8       \\
 \rule{0pt}{9pt}DSRG~(CVPR18)~\cite{huang2018weakly}  & VGG16 & I+S  & 59.0 & 60.4    \\
 \rule{0pt}{9pt}MCOF~(CVPR18)~\cite{wang2018weakly}  & VGG16 & I+S   & 56.2  & 57.6          \\
 \rule{0pt}{9pt}SeeNet~(NeurIPS)~\cite{hou2018self}  & VGG16 & I+S  & 61.1   & 60.7  \\
 \rule{0pt}{9pt}MDC~(CVPR18)~\cite{wei2018revisiting} & VGG16  & I+S   & 60.4    & 60.8    \\
 \rule{0pt}{9pt}FickleNet~(CVPR19)~\cite{lee2019ficklenet} & VGG16 & I+S  & 61.2    & 61.9  \\
 \rule{0pt}{9pt}OAA~(ICCV19)~\cite{jiang2019integral} & VGG16  & I+S     & 63.1   & 62.8     \\
  \rule{0pt}{9pt}MCIS~(ECCV20)~\cite{sun2020mining} & VGG16  & I+S   & 63.5   & 63.6    \\
 \rule{0pt}{9pt}ICD~(CVPR20)~\cite{fan2020learning} & VGG16 & I+S     & 64.0  & 63.9    \\
 \rule{0pt}{9pt}SGAN~(IEEEACC20)~\cite{yao2020saliency} & VGG16 & I+S    & 64.2  & 65.0   \\
 \rule{0pt}{9pt}DRS~(AAAI21)~\cite{kim2021discriminative}  & VGG16  & I+S   & 63.6  & 64.4   \\
 \rule{0pt}{9pt}NSROM~(CVPR21)~\cite{yao2021non}  & VGG16 & I+S   & 65.5   & 65.3   \\
  \rule{0pt}{9pt}EPS~(CVPR21)~\cite{lee2021railroad} & VGG16  & I+S   & 67.0   & 67.3   \\
\rule{0pt}{9pt}$I^{2}$CRC~(TMM22)~\cite{chen2022saliency} & VGG16  & I+S   & 64.3   & 65.4   \\
 \rule{0pt}{9pt}\textbf{MECPformer~(Ours)} & VGG16  & I+S  & \textbf{67.7} & \textbf{67.4}  \\ 
 \bottomrule
\end{tabular*}
\end{minipage}
\end{center}
\end{table}
\begin{table}[htbp]
\begin{center}
\begin{minipage}{260pt}
\caption{Quantitative comparisons to previous state-of-the-art approaches with ResNet101 backbone in segmentation network in the self-training stage. The result is with CRF}
\label{Tab.sota_resnet}
\begin{tabular*}{\textwidth}{@{\extracolsep{\fill}}lcccc@{\extracolsep{\fill}}}
\toprule
 \rule{0pt}{9pt}\textbf{Methods}  & \textbf{Backbone} & \textbf{Sup.} & \textbf{Val}  & \textbf{Test}  \\ 
 \midrule
 \rule{0pt}{9pt}SEAM~(CVPR20)~\cite{wang2020self}  & ResNet38  & I   & 64.5   & 65.7  \\
 \rule{0pt}{9pt}BES~(ECCV20)~\cite{chen2020weakly} & ResNet101  & I   & 65.7   & 66.6     \\
 \rule{0pt}{9pt}SC-CAM~(CVPR20)~\cite{chang2020weakly} & ResNet101  & I  & 66.1   & 65.9  \\
 \rule{0pt}{9pt}RRM~(AAAI20)~\cite{zhang2020reliability} & ResNet101 & I   & 66.3   & 66.5  \\
 \rule{0pt}{9pt}CONTA~(NeurIPS20)~\cite{zhang2020causal} & ResNet38 & I   & 66.1    & 66.7  \\
 \rule{0pt}{9pt}ECS-Net~(ICCV21)~\cite{sun2021ecs}  & ResNet38 & I   & 66.6   & 67.6 \\
 \rule{0pt}{9pt}CPN~(ICCV21)~\cite{zhang2021complementary}  & ResNet38 & I   & 67.8   & 68.5 \\
 \rule{0pt}{9pt}AdvCAM~(CVPR21)~\cite{lee2021anti}  & ResNet101 & I   & 68.1   & 68.0 \\
 \rule{0pt}{9pt}CGNet~(ICCV21)~\cite{kweon2021unlocking}  & ResNet38 & I   & 68.4   & 68.2 \\ 
  \rule{0pt}{9pt}SPML~(ICLR21)~\cite{ke2021universal}  & ResNet101 & I   & 69.5   & 71.6 \\ 
 \rule{0pt}{9pt}AFA~(CVPR22)~\cite{ru2022learning}  & MiT-B1 & I   & 66.0   & 66.3 \\ 
 \rule{0pt}{9pt}AMR~(AAAI22)~\cite{qin2022activation}  & ResNet50 & I   & 68.8   & 69.1 \\
 \rule{0pt}{9pt}SIPE~(CVPR22)~\cite{chen2022self}  & ResNet101 & I   & 68.8   & 69.7 \\ 
   \rule{0pt}{9pt}AD-CAM~(Arxiv22)~\cite{huang2022attention}  & ResNet101 & I & 69.0 & 69.4\\
  \rule{0pt}{9pt}ICAM~(CVPR22)~\cite{lee2022threshold}  & ResNet101 & I   & 70.7   & 70.6 \\
\midrule
  \rule{0pt}{9pt}SGAN~(IEEEACC20)~\cite{yao2020saliency} & ResNet101 & I+S   & 67.1  & 67.2 \\
 \rule{0pt}{9pt}CIAN~(AAAI20)~\cite{fan2020cian} & ResNet101 & I+S   & 64.3   & 65.3  \\
 \rule{0pt}{9pt}ICD~(CVPR20)~\cite{fan2020learning} & ResNet101  & I+S   &   67.8   & 68.0    \\
 \rule{0pt}{9pt}MCIS~(ECCV20)~\cite{sun2020mining} & ResNet101  & I+S   & 66.2    & 66.9  \\
 \rule{0pt}{9pt}AuxSegNet~(ICCV21)~\cite{xu2021leveraging} & ResNet38 & I+S   & 69.0   & 68.6 \\
  \rule{0pt}{9pt}{RIB}~(NeurIPS21)~\cite{lee2021reducing} & ResNet101 & I+S   & 70.2& 70.0 \\
 \rule{0pt}{9pt}{NSROM}~(CVPR21)~\cite{yao2021non} & ResNet101 & I+S   & 70.4   & 70.2 \\
 \rule{0pt}{9pt}{EPS}~(CVPR21)~\cite{lee2021railroad} & ResNet101 & I+S   & 70.9   & 70.8 \\
 \rule{0pt}{9pt}{EDAM}~(CVPR21)~\cite{wu2021embedded} & ResNet101 & I+S   & 70.9   & 70.6 \\
 \rule{0pt}{9pt}{DRS}~(AAAI21)~\cite{kim2021discriminative} & ResNet101 & I+S & 71.2& 71.4\\
 \rule{0pt}{9pt}$I^{2}$CRC~(TMM22)~\cite{chen2022saliency} & ResNet101  & I+S   & 69.3   & 69.5   \\
   \rule{0pt}{9pt}{Infer-CAM}~(WACV22)~\cite{sun2022inferring} & ResNet101 & I+S   & 70.8   & 71.8 \\
 \rule{0pt}{9pt}\textbf{MECPformer~(Ours)} & ResNet101 & I+S &\textbf{72.0}& \textbf{72.0}  \\
 \bottomrule
\end{tabular*}
\end{minipage}
\end{center}
\end{table}
Table \ref{Tab.sota_resnet} depicts the results of training on ResNet101. 
Aided by the saliency map, our MECPformer approach reached $72.0\%$ and $72.0\%$ on the validation and test sets, respectively. 
With the ResNet101 backbone, we also exceed the performance of NSROM \cite{yao2021non} and DRS \cite{kim2021discriminative} by $1.6\%$ and $0.8\%$. Furthermore, when it comes to EPS~\cite{lee2021railroad} and $I^{2}CRC$ \cite{chen2022saliency}, the results are boosted by $1.1\%$ and $2.7\%$.
The remarkable performance of our MECPformer on both the VGG16 and ResNet101 backbone confirms its effectiveness, yielding dense and trusted pseudo labels.
\begin{table}
\sidewaystablefn%
\begin{center}
\begin{minipage}{\textwidth}
\caption{Semantic segmentation performance on the PASCAL VOC 2012 validation set. The bottom group contains results with CRF refinement, while the top group is without CRF. Note that 17/21 classes obtain improvement using our approach w/o CRF. The best results are bolded}
\label{Tab.percls}
\resizebox{335pt}{!}{
\begin{tabular}{@{}l|ccccccccccccccccccccc|c@{}}
\toprule
\multicolumn{1}{l}{\rule{0pt}{9pt}Method}      & bkg           & aero          & bike          & bird          & boat         & bottle        & bus           & car           & cat           & chair         & cow           & table         & dog           & horse         & motor         & person        & plant         & sheep         & sofa          & train         & \multicolumn{1}{c}{tv}       & mIoU          \\ \hline
\multicolumn{1}{l}{\rule{0pt}{9pt}NSROM(w/o CRF)~\cite{yao2021non}}   & 91.2 & 85.4  & 38.2         & 75.3          & 65.0 & 73.7  & 89.3 & 76.8 & 88.0 & \textbf{25.7}          & \textbf{85.1} & \textbf{35.3}          & 81.1 & 80.8 & 75.8        & 78.6          & 51.1 & 81.2 & \textbf{38.0}         & 77.3 & \multicolumn{1}{c}{58.4}    & 69.1\\

\multicolumn{1}{l}{\rule{0pt}{9pt}\textbf{Ours(w/o CRF)}}    &      \textbf{92.2} & \textbf{87.9}          & \textbf{40.4}       
& \textbf{80.4}     & \textbf{73.3} & \textbf{75.3}        & \textbf{92.3} & \textbf{81.2} & \textbf{89.4} & 23.8          & 84.5 & 18.9          & \textbf{85.7} & \textbf{84.5} & \textbf{77.8}        & \textbf{81.9}         & \textbf{57.2} & \textbf{86.2} & 37.0         & \textbf{85.4} & \multicolumn{1}{c}{\textbf{60.9}}    & \textbf{71.3}\\
\midrule
\multicolumn{1}{l}{\rule{0pt}{9pt}MCOF~\cite{wang2018weakly}}         & 87.0          & 78.4          & 29.4          & 68.0          & 44.0          & 67.3          & 80.3          & 74.1          & 82.2          & 21.1          & 70.7          & 28.2          & 73.2          & 71.5          & 67.2          & 53.0          & 47.7          & 74.5          & 32.4          & 71.0          & \multicolumn{1}{l}{45.8}          & 60.3          \\
\multicolumn{1}{l}{\rule{0pt}{9pt}AffinityNet~\cite{ahn2018learning}}  & 88.2          & 68.2          & 30.6          & 81.1          & 49.6          & 61.0          & 77.8          & 66.1          & 75.1          & 29.0          & 66.0          & 40.2          & 80.4          & 62.0          & 70.4          & 73.7          & 42.5          & 70.7          & 42.6          & 68.1          & \multicolumn{1}{l}{51.6}          & 61.7          \\
\multicolumn{1}{l}{\rule{0pt}{9pt}Zeng et al.~\cite{zeng2019joint}}  & 90.0          & 77.4          & 37.5          & 80.7          & 61.6          & 67.9          & 81.8          & 69.0          & 83.7          & 13.6          & 79.4          & 23.3          & 78.0          & 75.3          & 71.4          & 68.1          & 35.2          & 78.2          & 32.5          & 75.5          & \multicolumn{1}{l}{48.0}          & 63.3          \\
\multicolumn{1}{l}{\rule{0pt}{9pt}SEAM~\cite{wang2020self}}         & 88.8          & 68.5          & 33.3          & \textbf{85.7} & 40.4          & 67.3          & 78.9          & 76.3          & 81.9          & 29.1          & 75.5          & 48.1          & 79.9          & 73.8          & 71.4          & 75.2          & 48.9          & 79.8          & 40.9          & 58.2          & \multicolumn{1}{c}{53.0}          & 64.5          \\
\multicolumn{1}{l}{\rule{0pt}{9pt}FickleNet~\cite{lee2019ficklenet}}    & 89.5          & 76.6          & 32.6          & 74.6          & 51.5          & 71.1          & 83.4          & 74.4          & 83.6          & 24.1          & 73.4          & 47.4          & 78.2          & 74.0          & 68.8          & 73.2          & 47.8          & 79.9          & 37.0          & 57.3          & \multicolumn{1}{c}{\textbf{64.6}} & 64.9          \\
\multicolumn{1}{l}{\rule{0pt}{9pt}SC-CAM~\cite{chang2020weakly}} & 88.8          & 51.6          & 30.3          & 82.9          & 53.0          & 75.8   & 88.6          & 74.8          & 86.6          & \textbf{32.4} & 79.9          & \textbf{53.8} & 82.3          & 78.5          & 70.4          & 71.2          & 40.2          & 78.3          & \textbf{42.9} & 66.8          & \multicolumn{1}{c}{58.8}          & 66.1          \\
\multicolumn{1}{l}{\rule{0pt}{9pt}NSROM(w/ CRF)~\cite{yao2021non}}   & 91.7 & 87.6  & 39.3          & 76.6          & 66.8 & 74.5  & 90.5 & 77.4 & 89.6 & 27.0          & \textbf{87.1} & 35.7          & 82.9 & 83.0 & 76.5        & 80.1          & 52.8 & 83.3 & 38.9          & 78.4 & \multicolumn{1}{c}{59.2}    & 70.4\\
\multicolumn{1}{l}{\rule{0pt}{9pt}\textbf{Ours(w/ CRF)}}    & \textbf{92.4} & \textbf{89.6}          & \textbf{41.1}          & 81.2         & \textbf{75.7} & \textbf{76.2}          & \textbf{93.1} & \textbf{81.5} & \textbf{90.6} & 23.5          & 85.7 & 17.6         & \textbf{87.2} & \textbf{85.3} & \textbf{78.1}        & \textbf{82.7}         & \textbf{57.5} & \textbf{87.5} & 37.6        & \textbf{86.1} & \multicolumn{1}{c}{62.6}                              & \textbf{72.0}\\ \bottomrule
\end{tabular}
}
\end{minipage}
\end{center}
\end{table}
In Table~\ref{Tab.percls}, we display detailed results for each individual category on the validation set. The two sets of results with (bottom) or without (top) CRF~\cite{krahenbuhl2011efficient} are compared. Our results show an effective improvement in 16 categories (including the background category) without CRF and superior to NSROM \cite{yao2021non} by $2.2\%$. It is worth noting that our approach without CRF is $0.9\%$ ahead of NSROM \cite{yao2021non} with CRF.
\begin{figure*}[!t]
    \centering
     \includegraphics[scale=0.4]{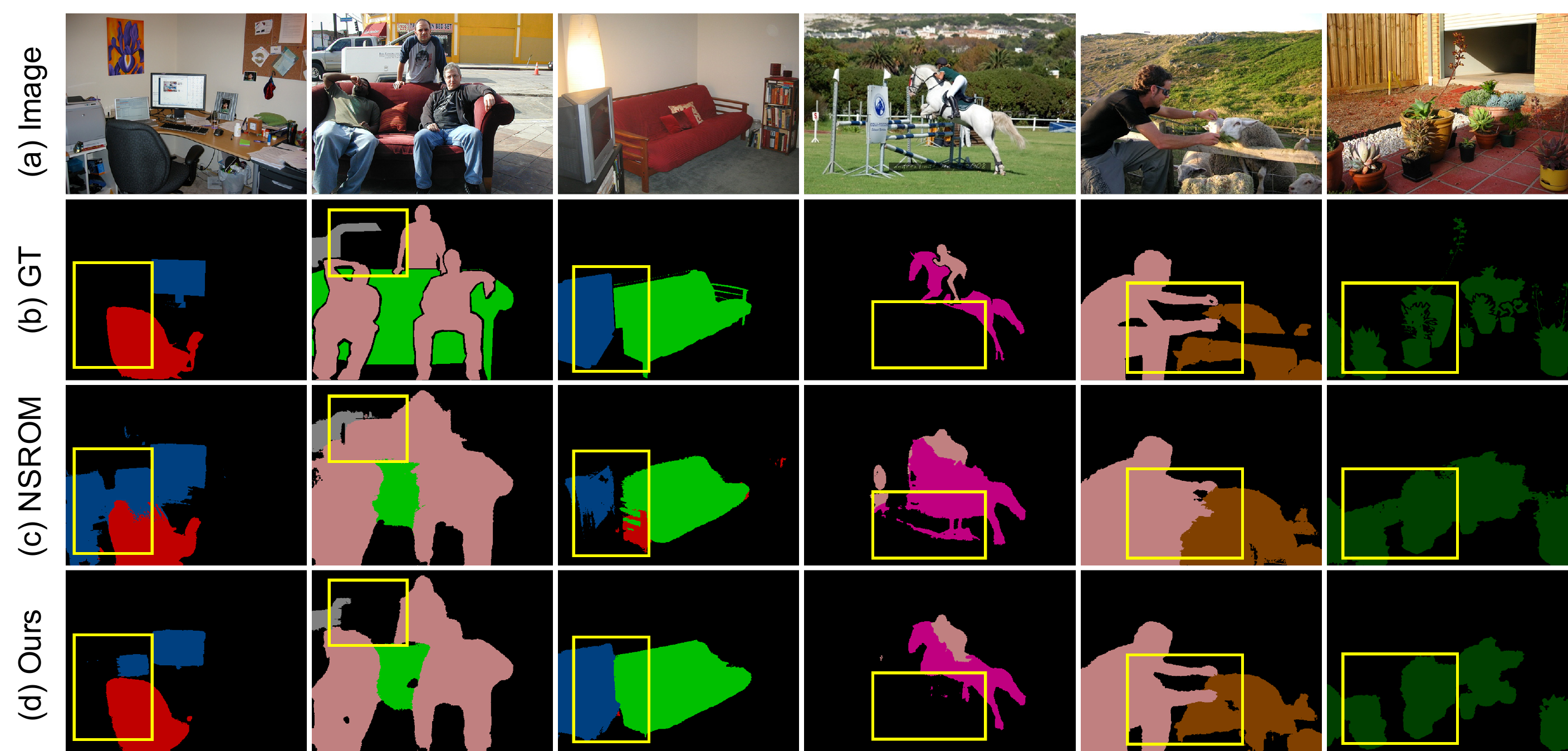}
    \caption{Qualitative segmentation results on the PASCAL VOC 2012 validation set. (a) Input image, (b) Ground truth, (c) Prediction of NSROM~\protect\cite{yao2021non}, (d) Prediction of Ours. We can find that our approach demonstrates finer-grained boundary information. Best viewed in color}
    \label{Fig.finalresult}
\end{figure*}
Fig. \ref{Fig.finalresult} visualizes the qualitative segmentation results on PASCAL VOC 2012 validation set. From the reported images, the pseudo labels arising from our method have delicate borders and successfully mitigate the false positive flaws. 

\subsubsection{MS COCO}
\begin{table}[ht]
\begin{center}
\begin{minipage}{260pt}
\caption{Performance comparison of WSSS methods in terms of mIoU (\%) on the MS COCO validation set.}
\label{Tab.coco}
\begin{tabular*}{\textwidth}{@{\extracolsep{\fill}}lccc@{\extracolsep{\fill}}}
\toprule
 \rule{0pt}{9pt} \textbf{Method}&\textbf{Backbone}&\textbf{Sup.}&\textbf{Val}\\
\midrule
 \rule{0pt}{9pt}  \text{Wang et al.~(IJCV20)~\cite{wang2020weakly}} & \text{VGG16} & \text{I} &  \text{27.7}\\
 \rule{0pt}{9pt} \text{Luo et al.~(AAAI20)~\cite{luo2020learning}}  & \text{VGG16} & \text{I} & \text{29.9}\\
 \rule{0pt}{9pt} \text{SEAM~(CVPR20)~\cite{wang2020self}} & \text{ResNet38} & \text{I} & \text{31.9}\\
 \rule{0pt}{9pt}  \text{CONTA (NeurIPS20)~\cite{zhang2020causal}} & \text{ResNet101} & \text{I} & \text{33.4}\\
  \rule{0pt}{9pt} \text{CDA(ICCV21)~\cite{su2021context}} &\text{ResNet38} &\text{I} & \text{33.2}\\
   \rule{0pt}{9pt} \text{SLRNet(IJCV22)~\cite{pan2022learning}} &\text{ResNet38} &\text{I} & \text{35.0}\\
 \midrule
  \rule{0pt}{9pt} \text{DSRG(CVPR18)~\cite{huang2018weakly}}&\text{VGG16}& \text{I+S} &  \text{26.0}\\
  \rule{0pt}{9pt} \text{$I^{2}$CRC~(TMM22)~\cite{chen2022saliency}}&\text{VGG16}& \text{I+S} &  \text{31.2}\\
  \rule{0pt}{9pt} \text{AuxSegNet~(ICCV21)\cite{lee2021reducing}} & \text{ResNet38} & \text{I+S} &  \text{33.9}\\
    \rule{0pt}{9pt} \text{G-WSSS(AAAI21)~\cite{zhou2021group}}&\text{ResNet101}& \text{I+S} &  \text{28.4}\\
 \rule{0pt}{9pt} \text{EPS~(CVPR21)\cite{lee2021railroad}} & \text{ResNet101} & \text{I+S} & \text{35.7}\\
 \rule{0pt}{9pt} \textbf{MECPformer~(Ours)} & \text{ResNet101} & \text{I+S} &  \textbf{42.4}\\
 \bottomrule
\end{tabular*}
\end{minipage}
\end{center}
\end{table}
\begin{figure*}[!t]
    \centering
     \includegraphics[scale=0.4]{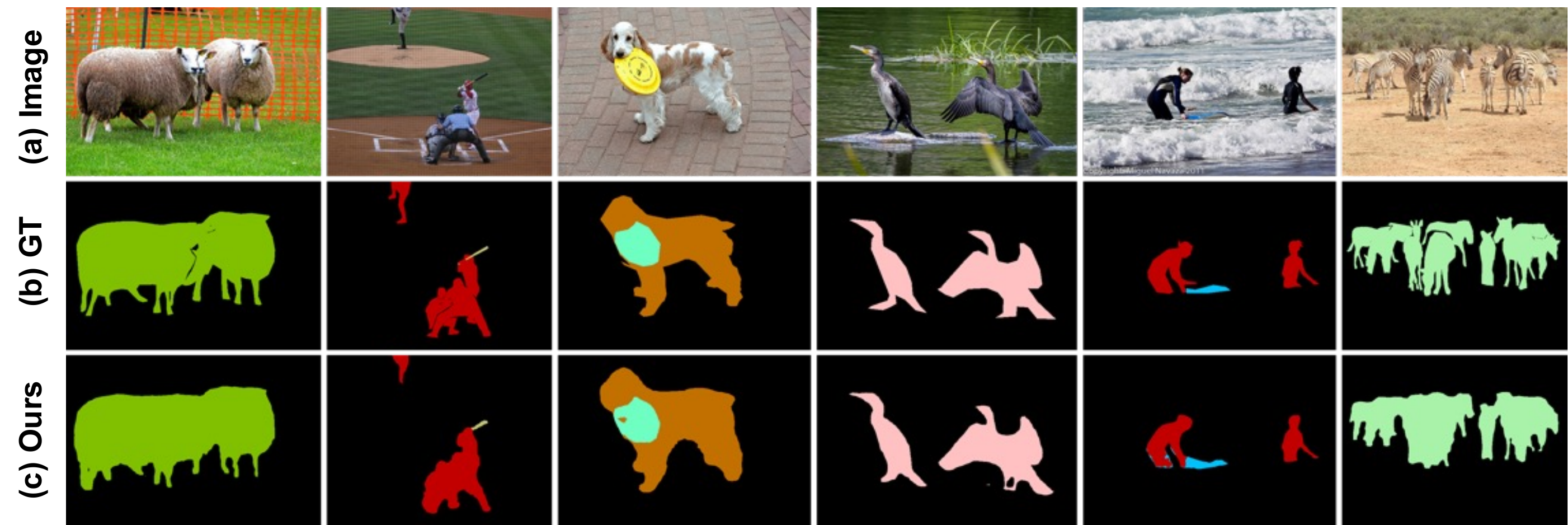}
    \caption{Qualitative segmentation results on the MS COCO 2014 validation set. (a) Input image, (b) Ground truth, (c) Prediction of Ours. Best viewed in color}
    \label{Fig.coco}
\end{figure*}
Table \ref{Tab.coco} displays the performance results obtained on the MS COCO dataset. Our method outperforms EPS by $6.3\%$ and surpasses the image-level label-only approach by $7.0\%$. These outcomes suggest that our method performs exceptionally well on even the more challenging MS COCO dataset and has great potential. The qualitative results on MS COCO are showcased in Fig. \ref{Fig.coco}.

\subsection{Ablation Studies}
In this chapter, we analyze a series of ablation studies to evaluate the effectiveness of our designed method's different modules and parameter settings. 

\subsubsection{Effect of Component}
\begin{table}[ht]
\begin{center}
\begin{minipage}{260pt}
\caption{Ablation study of each component in MECPformer. ``MECP'' means Multi-estimations Complementary Patch strategy. ``Sal'' denotes the saliency map. ``ACM'' is the Adaptation Conflict Module.
The mIoU result of the validation set on the PASCAL VOC 2012 containing self-training and CRF}
\label{Tab.sumAblation}
\begin{tabular*}{\textwidth}{@{\extracolsep{\fill}}lcccc@{\extracolsep{\fill}}}
\toprule
 \rule{0pt}{9pt} \textbf{MECP}&\textbf{Sal}&\textbf{ACM}&\textbf{CRF}&\textbf{mIoU}\\
\midrule
 \rule{0pt}{9pt} \text{ } & \text{ } & \text{ } & \text{ } &  \text{62.2}\\
 \rule{0pt}{9pt} \text{ } & \text{ } & \text{ } & \checkmark &  \text{65.2}\\
 \rule{0pt}{9pt} \checkmark & \text{ } & \text{ } & \text{ } &  \text{64.0}\\
 \rule{0pt}{9pt} \checkmark & \text{ } & \text{ } & \checkmark &  \text{66.8}\\
 \rule{0pt}{9pt} \checkmark & \checkmark & \text{ } & \text{ } & \text{67.7}\\
 \rule{0pt}{9pt} \checkmark & \checkmark & \text{ } & \checkmark & \text{68.2}\\
 \rule{0pt}{9pt} \checkmark & \checkmark & \checkmark & \text{ } & \text{71.3}\\
 \rule{0pt}{9pt} \checkmark & \checkmark & \checkmark & \checkmark & \text{72.0}\\ \bottomrule
\end{tabular*}
\end{minipage}
\end{center}
\end{table}
We perform ablation studies to confirm the validity of our proposed main contributions, which include the Multi-estimations Complementary Patch (MECP) strategy and Adaptation Conflict Module (ACM). We further demonstrate the effect of the saliency map and CRF on final results, and the experimental analysis is shown in Table ~\ref{Tab.sumAblation}. The results are derived after self-training of the segmentation network and CRF on the validation set.

The initial result of $62.2\%$ is generated by TransCAM \cite{li2022transcam}, and our MECP strategy boosts $1.8\%$. In addition,
We observe that the saliency map contributes about $3.7\%$ boost without CRF post-processing. When it comes to ACM, there is a $3.6\%$ and $3.8\%$ increment without and with CRF, respectively. Each of the key components contributes prominently to the final result. With the assistance of MECP, saliency map, and ACM, we achieved $9.1\%$ improvement without CPF.

\subsubsection{Multi-estimations $k$ in MECP}
Table~\ref{Tab.MECPk} illustrates mIoU of the initial seed on the PASCAL VOC train set, consisting of 1464 images. The values are derived in the vicinity of the CPN~\cite{zhang2021complementary} best result, where $k$ equals 300. In order to obtain 5 different values of $k$, corresponding to epochs that are divisible by 0, 1, 2, 3, and 4, respectively, we conducted experiments. The table displays 8 different sets of $k$ values and it can be observed that the optimal initial seed is achieved when $k$ is selected as (200, 250, 300, 350, 400).
\begin{table}[ht]
\begin{center}
\begin{minipage}{330pt}
\caption{Ablation study of multi-estimations $k$ in MECP. We show the initial seed mIoU results on the PASCAL VOC train set.}
\label{Tab.MECPk}
\begin{tabular*}{\textwidth}{@{\extracolsep{\fill}}lcclcc@{\extracolsep{\fill}}}
\toprule
 \rule{0pt}{9pt}\textbf{No.}&\textbf{k}&\textbf{mIoU}&\textbf{No.}&\textbf{k}&\textbf{mIoU}\\
\midrule
 \rule{0pt}{9pt} \text{1} & \text{(50, 100, 150, 200, 250)} & \text{62.1} & \text{5} &  \text{(50, 150, 250, 250, 450)}& \text{65.3}\\
 \rule{0pt}{9pt} \text{2} & \text{(100, 150, 200, 250, 300)} & \text{63.4} & \text{6} &  \text{(100, 200, 300, 400, 500)}& \text{64.8}\\
 \rule{0pt}{9pt} \text{3} & \text{(150, 200, 250, 300, 350)} & \text{63.7} & \text{7}&  \text{(200, 250, 300, 350, 400)} &  \textbf{66.6}\\
 \rule{0pt}{9pt} \text{4} & \text{(250, 300, 350, 400, 450)} & \text{64.0} & \text{8}& \text{(300, 350, 400, 450, 500)} &  \text{65.2}\\
\bottomrule
\end{tabular*}
\end{minipage}
\end{center}
\end{table}

\subsubsection{Conflict Rate in ACM}
\begin{figure}[htbp]
    \centering
    \scalebox{1}{\includegraphics[scale=0.3]{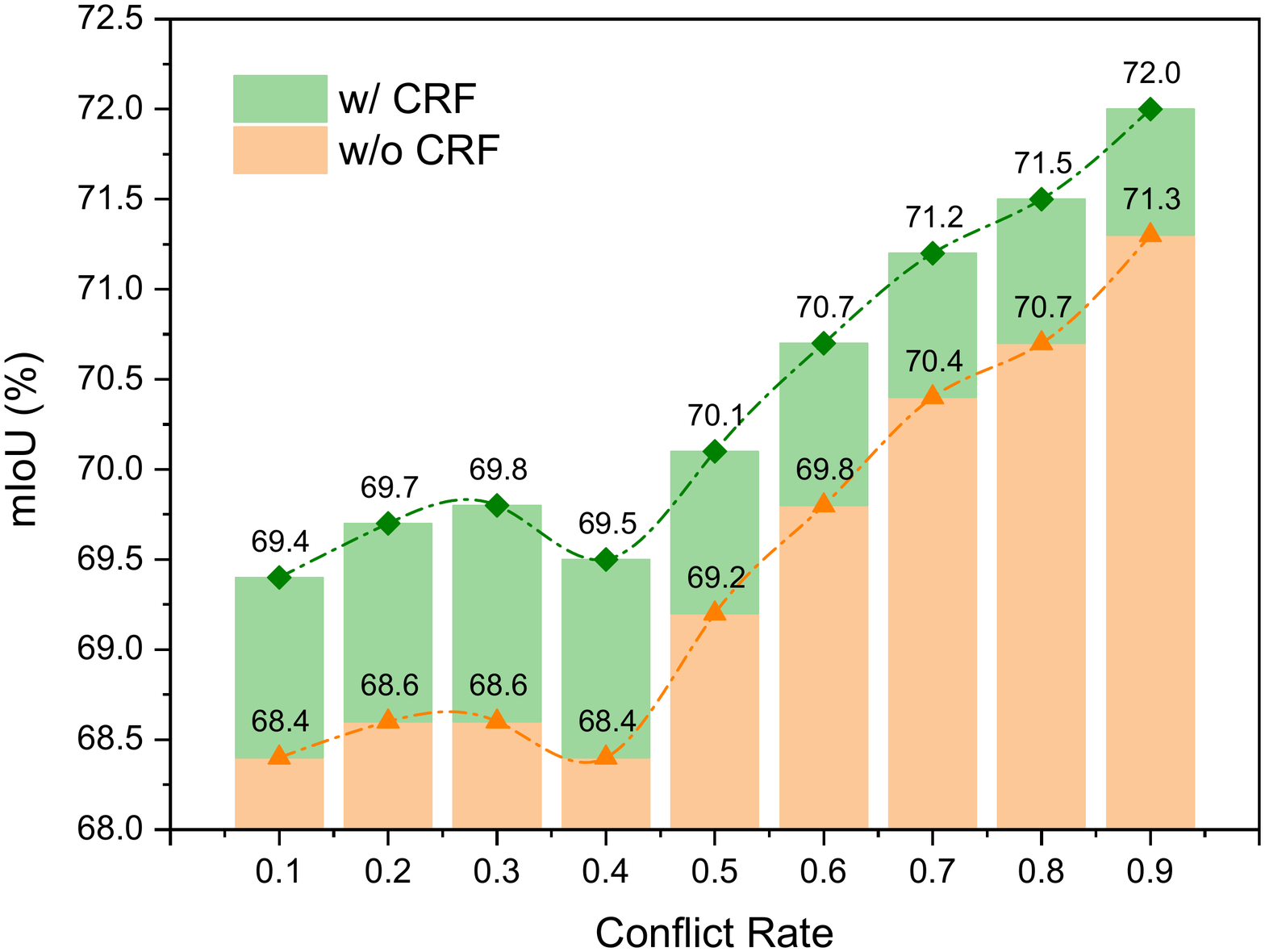}}
    \caption{Conflict rate analysis in Adaptation Conflict Module (ACM). The best result of $71.3\%$ is obtained when the conflict rate equals $0.9$ without CRF. We can also observe that when the conflict rate decreases, the ACM mistakes a large amount of valid information for conflicting information, resulting in a drop in the worst result of $68.4\%$, a difference of $2.9\%$ points. As shown in green, we present the mIoU values with CRF and the growth trend of the data remains consistent with that observed before applying CRF.
    }
    \label{Fig.Cr}
\end{figure}
We perform ablation studies of conflict rates in ACM in Fig. \ref{Fig.Cr}. We report the mIoU results on the PASCAL VOC 2012 validation set after self-training without CRF, as shown in the yellow part of Fig. \ref{Fig.Cr}. The conflict rate ranges from 0 to 1, taking the value every $0.1$. The best segmentation results are obtained when the conflict rate equals $0.9$. 
In addition, It is worth noting that the quality of the pseudo labels degrades when the conflict rate is gradually decreasing. The reason is that ACM considers a large number of pixels as conflicting pixels and marks them as $255$, leading to a reduction in the range of plausible pseudo labels.
As can be seen, the best result $71.3\%$ outperforms the worst result $68.4\%$ by $2.9\%$.
Furthermore, we demonstrate the mIoU post CRF, as shown in green in Fig. \ref{Fig.Cr}. Our results indicate that the CRF post-processing technique yields an approximate improvement of $1\%$ in performance. Notably, the optimal mIoU value of $72.0\%$ is attained at a conflict rate of $0.9$, deviating by $2.6\%$ from the minimum value, while the growth trend of the data remains consistent with that observed before applying CRF.

\begin{figure}[htbp]
    \centering
    \scalebox{1}{\includegraphics[scale=0.45]{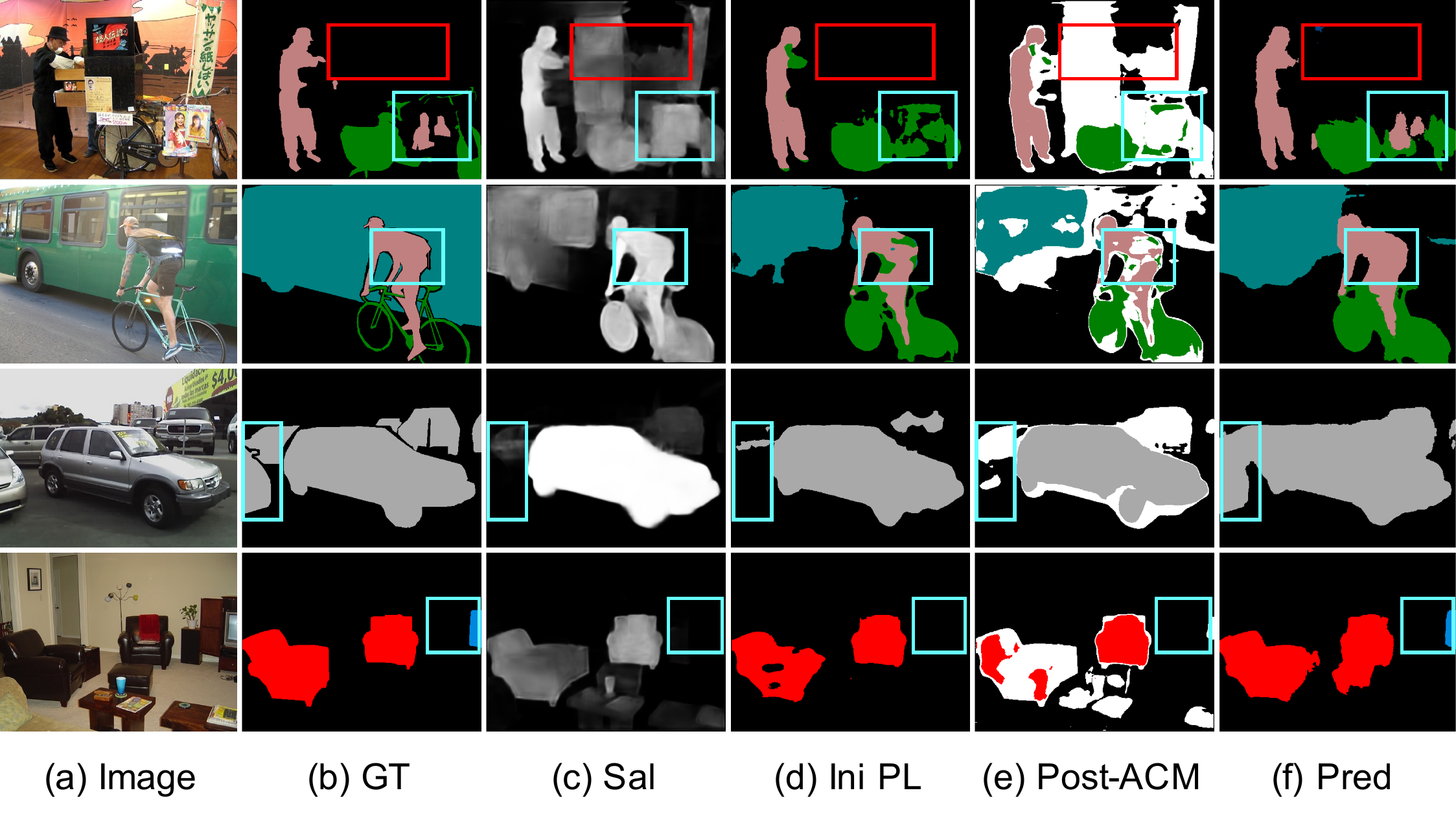}}
    \caption{Visualization of ACM processing effects. The ground truth and saliency map are indicated by ``GT'' and ``Sal'', respectively. ``Ini PL'' refers to the initial pseudo label after a simple threshold and saliency map. ``Post-ACM'' denotes the Ini PL after ACM module processing. ``Pred'' is the final prediction results.
    As shown in the first two rows, we observe from columns (d) and (e) that ACM can check for conflicting pixels and obtain more accurate predictions with the networks' self-learning capability. As seen from the last two rows, ACM is sufficiently capable of mining the target outside the saliency map
    }
    \label{Fig.vis_acm}
\end{figure}
\subsubsection{Visualization of ACM Results}
To demonstrate the mechanism action of ACM more visually, we present the results in Fig. \ref{Fig.vis_acm}.
``GT'' denotes ground truth. ``Sal'' refers to the saliency map. ``Ini PL'' is the initial pseudo label for extracting background information via the saliency map and simple threshold. ``Ini PL'' further manipulates by ACM to yield ``Post-ACM''. ``Pred'' indicates the final prediction result. From the results of the first two rows, ACM identifies the potential conflicting pixels and labels them as 255 (see (d) and (e)). The segmentation network's self-learning capability is then exploited to gain the correct prediction results. The red box indicates that the saliency map considers a conflicting pixel belonging to the valid class. Still, the initial pseudo-label considers it the background class, and the segmentation network can circumvent it well. In addition, since the saliency map is class agnostic and not always trustworthy (compare (b) and (c)), ACM enables excavating the regions outside the saliency map (presented in the last two rows).

\subsubsection{Results on Different Backbone}
\begin{table}[ht]
\begin{center}
\begin{minipage}{260pt}
\caption{Analysis of different segmentation frameworks. More robust segmentation frameworks are essential for WSSS. The backbone is pre-trained on ImageNet.}
\label{Tab.seg}
\begin{tabular*}{\textwidth}{@{\extracolsep{\fill}}lccc@{\extracolsep{\fill}}}
\toprule
 \rule{0pt}{9pt}   \textbf{Network} & \textbf{Sup.} & \textbf{val} & \textbf{test}\\
\midrule
 \rule{0pt}{9pt}   \text{DeepLab-V2} & I+S & \text{70.7}  & \text{70.5}\\
\midrule
 \rule{0pt}{9pt}   \text{DeepLab-V3+} & I+S & \text{71.0}  & \text{71.9}\\
\bottomrule
\end{tabular*}
\end{minipage}
\end{center}
\end{table}

We present the segmentation results on different backbones adopting ImageNet pre-training weights for a fairer comparison. The results are post-CRF. Table \ref{Tab.seg} demonstrates the final segmentation results on DeepLab-v2/v3+. We remark that the network is stronger, and the final result is superior.

\subsection{Discussions}
\begin{figure}[htbp]
    \centering
     \includegraphics[scale=0.4]{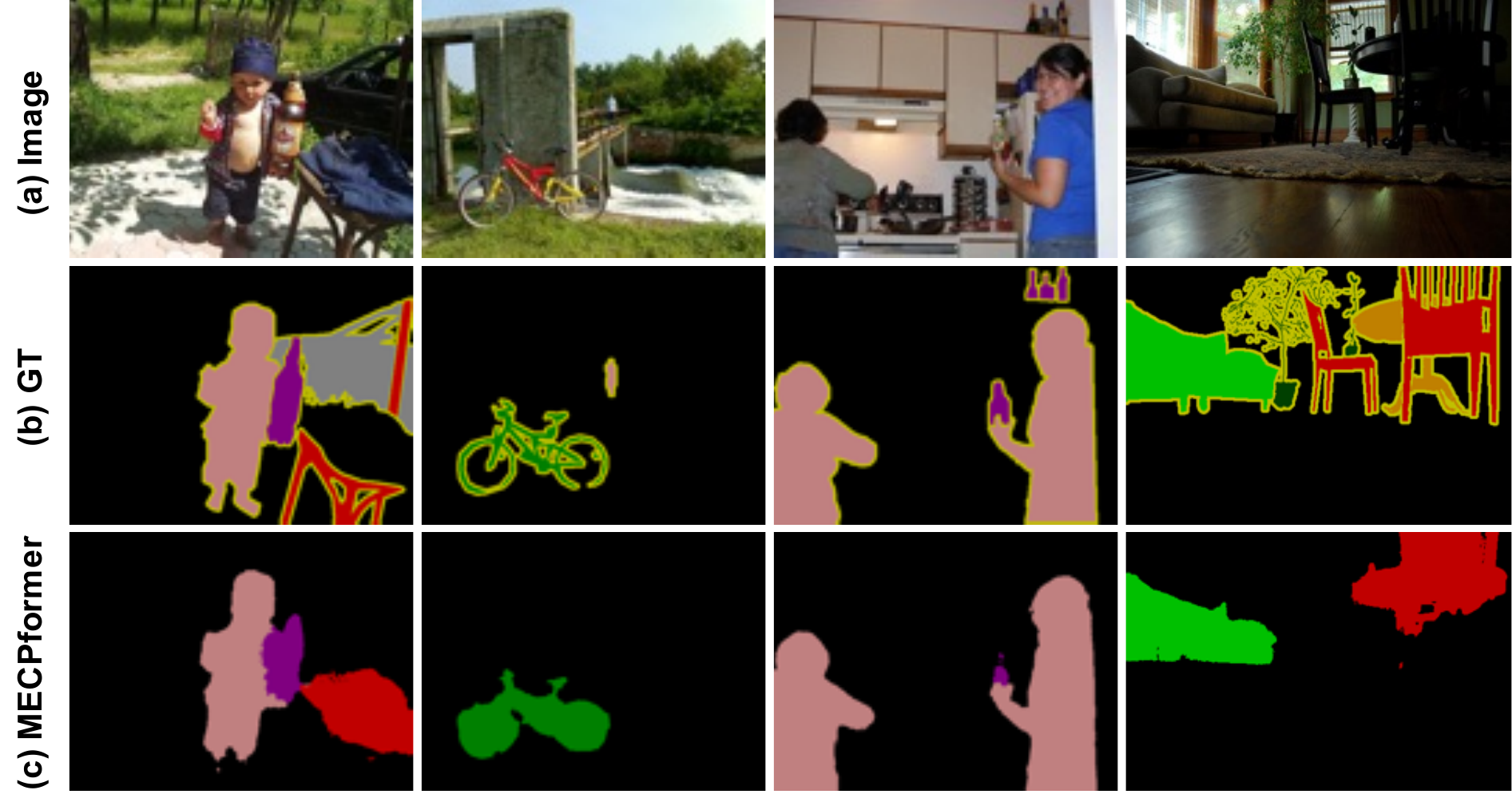}
    \caption{List of Failure cases. We list some bad cases to make readers better understand our work. Best viewed in color}
    \label{Fig.failure_cases}
\end{figure}

Our MECPformer performs effectively on most images, As illustrated in Fig. \ref{Fig.finalresult}. However, it inevitably suffers from some limitations, as shown in Fig. \ref{Fig.failure_cases}.
First, it might yield sub-optimal results when there are multiple categories in a single image and the classes are closely related, as indicated in the first column in Fig. \ref{Fig.failure_cases}. To analyze this phenomenon more clearly, we evaluate the segmentation results on the number of classes in a single image in Table \ref{Tab.dif_cls}. 
\begin{table}[htbp]
\begin{center}
\begin{minipage}{260pt}
\caption{The mIoU of images with the different numbers of categories on the validation set of PASCAL VOC 2012. Multi-semantic target images are still challenging.}
\label{Tab.dif_cls}
\begin{tabular*}{\textwidth}{@{\extracolsep{\fill}}lcccc@{\extracolsep{\fill}}}
\toprule
\multicolumn{1}{l}{\rule{0pt}{9pt} \textbf{Classes}}   & \textbf{One} & \textbf{Two} & \multicolumn{1}{c}{\textbf{Three+}} & \textbf{mIoU}\\
\midrule
\multicolumn{1}{l}{\rule{0pt}{9pt} \textbf{Val (w/o CRF)}}  & \text{76.1} & 
 \text{65.8}& \multicolumn{1}{c}{\text{55.6}}  & \text{71.3}\\
\multicolumn{1}{l}{\rule{0pt}{9pt} \textbf{Val (w/ CRF)}} & \text{77.1} & 
 \text{66.8}& \multicolumn{1}{c}{\text{55.5}}   & \text{72.0}\\
\bottomrule
\end{tabular*}
\end{minipage}
\end{center}
\end{table}
From Table \ref{Tab.dif_cls}, we find that the image segmentation quality gradually decreases as the number of categories in a single image grows, which indicates that weakly supervised image-level segmentation with multiple semantic categories is still a challenging task. The natural advantages of the Transformer bring fresh thinking towards the issue to some extent. 
In addition, when the target in the image is too tiny and blends in with the background, the target may be incorrectly identified as the background, as reflected in the second and third columns in Fig. \ref{Fig.failure_cases}. 
Third, the light in the picture is overexposed or too low, with possibly unsatisfactory results as well, as indicated in the last column of Fig. \ref{Fig.failure_cases},  Enhancing the network's learning power and introducing additional labels may alleviate these problems, which is the future direction we will focus on.

In addition, we have found that the employment of Transformer-based approaches in the segmentation stage shows greater potential \cite{chen2022semformer}. This finding has motivated us to introduce more robust segmentation stage networks into weakly supervised semantic segmentation, which will be a focus of our future endeavors.

Moreover, We obtained highly promising results during the initial seed phase, albeit with suboptimal segmentation outcomes compared to  MCTformer \cite{xu2022multi} due to the adoption of simplistic post-processing techniques. This suggests that further exploration into post-processing methodologies may lead to unexpected segmentation outcomes. Hence, our future work will focus on delving deeper into post-processing methods.

Finally, based on the Conformer-S approach, our method has less complexity compared to the ResNet38-based method, but there is still significant room for improvement compared to lightweight models. Further optimization of the attention mechanism and network model to gradually approach the lightweight model is another issue of our future focus.

\section{Conclusion}
In this work, we demonstrate MECPformer, equipped with Multi-estimations Complementary Patch (MECP) strategy and Adaptation Conflict Module (ACM). 
MECP strengthens the learning capability of the network, enabling the network to explore and merge different levels of semantic features, alleviating the false-positive defect. Subsequently, ACM removes the conflicting uncertain pixels and further resolves the false positive pixels by leveraging the self-learning capability of the segmentation network.
Extensive experiments demonstrate that our MECPformer has achieved superior results.
We hope MECPformer based on Transformers can demonstrate a novel perspective for weakly supervised semantic segmentation.

\backmatter

\section*{Declarations}

\begin{itemize}
    \item Funding
    
    The authors did not receive support from any organization for the submitted work.
    
    \item Conflicts of interests/Competing interests

    The authors have no relevant financial or non-financial interests to disclose.
    
    \item Availability of data and materials

    The datasets generated during and/or analysed during the current study are available in the GitHub repository, https://drive.google.com/file/d/1-2cYvmJ4u52FvZ722eIti1jf5hfGr6Rz/view

    \item Code availability
    
    Code and pre-trained models will be available online soon.
    
    \item Authors' contributions
    
    All authors contributed to the study conception and design. Material preparation, data collection and analysis were performed by Chunmeng Liu, Guangyao Li, Yao Shen, and Ruiqi Wang. The first draft of the manuscript was written by Chunmeng Liu and all authors commented on previous versions of the manuscript. All authors read and approved the final manuscript.
    \item Ethics approval (Not applicable)
    \item Consent to participate (Not applicable)
    \item Consent for publication (Not applicable)
    
\end{itemize}

\bibliography{sn-bibliography}


\begin{thebibliography}{77}
\ifx \bisbn   \undefined \def \bisbn  #1{ISBN #1}\fi
\ifx \binits  \undefined \def \binits#1{#1}\fi
\ifx \bauthor  \undefined \def \bauthor#1{#1}\fi
\ifx \batitle  \undefined \def \batitle#1{#1}\fi
\ifx \bjtitle  \undefined \def \bjtitle#1{#1}\fi
\ifx \bvolume  \undefined \def \bvolume#1{\textbf{#1}}\fi
\ifx \byear  \undefined \def \byear#1{#1}\fi
\ifx \bissue  \undefined \def \bissue#1{#1}\fi
\ifx \bfpage  \undefined \def \bfpage#1{#1}\fi
\ifx \blpage  \undefined \def \blpage #1{#1}\fi
\ifx \burl  \undefined \def \burl#1{\textsf{#1}}\fi
\ifx \doiurl  \undefined \def \doiurl#1{\url{https://doi.org/#1}}\fi
\ifx \betal  \undefined \def \betal{\textit{et al.}}\fi
\ifx \binstitute  \undefined \def \binstitute#1{#1}\fi
\ifx \binstitutionaled  \undefined \def \binstitutionaled#1{#1}\fi
\ifx \bctitle  \undefined \def \bctitle#1{#1}\fi
\ifx \beditor  \undefined \def \beditor#1{#1}\fi
\ifx \bpublisher  \undefined \def \bpublisher#1{#1}\fi
\ifx \bbtitle  \undefined \def \bbtitle#1{#1}\fi
\ifx \bedition  \undefined \def \bedition#1{#1}\fi
\ifx \bseriesno  \undefined \def \bseriesno#1{#1}\fi
\ifx \blocation  \undefined \def \blocation#1{#1}\fi
\ifx \bsertitle  \undefined \def \bsertitle#1{#1}\fi
\ifx \bsnm \undefined \def \bsnm#1{#1}\fi
\ifx \bsuffix \undefined \def \bsuffix#1{#1}\fi
\ifx \bparticle \undefined \def \bparticle#1{#1}\fi
\ifx \barticle \undefined \def \barticle#1{#1}\fi
\bibcommenthead
\ifx \bconfdate \undefined \def \bconfdate #1{#1}\fi
\ifx \botherref \undefined \def \botherref #1{#1}\fi
\ifx \url \undefined \def \url#1{\textsf{#1}}\fi
\ifx \bchapter \undefined \def \bchapter#1{#1}\fi
\ifx \bbook \undefined \def \bbook#1{#1}\fi
\ifx \bcomment \undefined \def \bcomment#1{#1}\fi
\ifx \oauthor \undefined \def \oauthor#1{#1}\fi
\ifx \citeauthoryear \undefined \def \citeauthoryear#1{#1}\fi
\ifx \endbibitem  \undefined \def \endbibitem {}\fi
\ifx \bconflocation  \undefined \def \bconflocation#1{#1}\fi
\ifx \arxivurl  \undefined \def \arxivurl#1{\textsf{#1}}\fi
\csname PreBibitemsHook\endcsname

\bibitem{feng2020deep}
\begin{barticle}
\bauthor{\bsnm{Feng}, \binits{D.}},
\bauthor{\bsnm{Haase-Sch{\"u}tz}, \binits{C.}},
\bauthor{\bsnm{Rosenbaum}, \binits{L.}},
\bauthor{\bsnm{Hertlein}, \binits{H.}},
\bauthor{\bsnm{Glaeser}, \binits{C.}},
\bauthor{\bsnm{Timm}, \binits{F.}},
\bauthor{\bsnm{Wiesbeck}, \binits{W.}},
\bauthor{\bsnm{Dietmayer}, \binits{K.}}:
\batitle{Deep multi-modal object detection and semantic segmentation for
  autonomous driving: Datasets, methods, and challenges}.
\bjtitle{IEEE Transactions on Intelligent Transportation Systems}
\bvolume{22}(\bissue{3}),
\bfpage{1341}--\blpage{1360}
(\byear{2020})
\end{barticle}
\endbibitem

\bibitem{weng2021stage}
\begin{botherref}
\oauthor{\bsnm{Weng}, \binits{X.}},
\oauthor{\bsnm{Yan}, \binits{Y.}},
\oauthor{\bsnm{Chen}, \binits{S.}},
\oauthor{\bsnm{Xue}, \binits{J.-H.}},
\oauthor{\bsnm{Wang}, \binits{H.}}:
Stage-aware feature alignment network for real-time semantic segmentation of
  street scenes.
IEEE Transactions on Circuits and Systems for Video Technology
(2021)
\end{botherref}
\endbibitem

\bibitem{lee2021bbam}
\begin{bchapter}
\bauthor{\bsnm{Lee}, \binits{J.}},
\bauthor{\bsnm{Yi}, \binits{J.}},
\bauthor{\bsnm{Shin}, \binits{C.}},
\bauthor{\bsnm{Yoon}, \binits{S.}}:
\bctitle{Bbam: Bounding box attribution map for weakly supervised semantic and
  instance segmentation}.
In: \bbtitle{Proceedings of the IEEE/CVF Conference on Computer Vision and
  Pattern Recognition},
pp. \bfpage{2643}--\blpage{2652}
(\byear{2021})
\end{bchapter}
\endbibitem

\bibitem{lin2016scribblesup}
\begin{bchapter}
\bauthor{\bsnm{Lin}, \binits{D.}},
\bauthor{\bsnm{Dai}, \binits{J.}},
\bauthor{\bsnm{Jia}, \binits{J.}},
\bauthor{\bsnm{He}, \binits{K.}},
\bauthor{\bsnm{Sun}, \binits{J.}}:
\bctitle{Scribblesup: Scribble-supervised convolutional networks for semantic
  segmentation}.
In: \bbtitle{Proceedings of the IEEE Conference on Computer Vision and Pattern
  Recognition},
pp. \bfpage{3159}--\blpage{3167}
(\byear{2016})
\end{bchapter}
\endbibitem

\bibitem{bearman2016s}
\begin{bchapter}
\bauthor{\bsnm{Bearman}, \binits{A.}},
\bauthor{\bsnm{Russakovsky}, \binits{O.}},
\bauthor{\bsnm{Ferrari}, \binits{V.}},
\bauthor{\bsnm{Fei-Fei}, \binits{L.}}:
\bctitle{What’s the point: Semantic segmentation with point supervision}.
In: \bbtitle{European Conference on Computer Vision},
pp. \bfpage{549}--\blpage{565}
(\byear{2016}).
\bcomment{Springer}
\end{bchapter}
\endbibitem

\bibitem{ahn2018learning}
\begin{bchapter}
\bauthor{\bsnm{Ahn}, \binits{J.}},
\bauthor{\bsnm{Kwak}, \binits{S.}}:
\bctitle{Learning pixel-level semantic affinity with image-level supervision
  for weakly supervised semantic segmentation}.
In: \bbtitle{Proceedings of the IEEE Conference on Computer Vision and Pattern
  Recognition},
pp. \bfpage{4981}--\blpage{4990}
(\byear{2018})
\end{bchapter}
\endbibitem

\bibitem{lee2021railroad}
\begin{bchapter}
\bauthor{\bsnm{Lee}, \binits{S.}},
\bauthor{\bsnm{Lee}, \binits{M.}},
\bauthor{\bsnm{Lee}, \binits{J.}},
\bauthor{\bsnm{Shim}, \binits{H.}}:
\bctitle{Railroad is not a train: Saliency as pseudo-pixel supervision for
  weakly supervised semantic segmentation}.
In: \bbtitle{Proceedings of the IEEE/CVF Conference on Computer Vision and
  Pattern Recognition},
pp. \bfpage{5495}--\blpage{5505}
(\byear{2021})
\end{bchapter}
\endbibitem

\bibitem{wu2021embedded}
\begin{bchapter}
\bauthor{\bsnm{Wu}, \binits{T.}},
\bauthor{\bsnm{Huang}, \binits{J.}},
\bauthor{\bsnm{Gao}, \binits{G.}},
\bauthor{\bsnm{Wei}, \binits{X.}},
\bauthor{\bsnm{Wei}, \binits{X.}},
\bauthor{\bsnm{Luo}, \binits{X.}},
\bauthor{\bsnm{Liu}, \binits{C.H.}}:
\bctitle{Embedded discriminative attention mechanism for weakly supervised
  semantic segmentation}.
In: \bbtitle{Proceedings of the IEEE/CVF Conference on Computer Vision and
  Pattern Recognition},
pp. \bfpage{16765}--\blpage{16774}
(\byear{2021})
\end{bchapter}
\endbibitem

\bibitem{wei2017object}
\begin{bchapter}
\bauthor{\bsnm{Wei}, \binits{Y.}},
\bauthor{\bsnm{Feng}, \binits{J.}},
\bauthor{\bsnm{Liang}, \binits{X.}},
\bauthor{\bsnm{Cheng}, \binits{M.-M.}},
\bauthor{\bsnm{Zhao}, \binits{Y.}},
\bauthor{\bsnm{Yan}, \binits{S.}}:
\bctitle{Object region mining with adversarial erasing: A simple classification
  to semantic segmentation approach}.
In: \bbtitle{Proceedings of the IEEE Conference on Computer Vision and Pattern
  Recognition},
pp. \bfpage{1568}--\blpage{1576}
(\byear{2017})
\end{bchapter}
\endbibitem

\bibitem{huang2018weakly}
\begin{bchapter}
\bauthor{\bsnm{Huang}, \binits{Z.}},
\bauthor{\bsnm{Wang}, \binits{X.}},
\bauthor{\bsnm{Wang}, \binits{J.}},
\bauthor{\bsnm{Liu}, \binits{W.}},
\bauthor{\bsnm{Wang}, \binits{J.}}:
\bctitle{Weakly-supervised semantic segmentation network with deep seeded
  region growing}.
In: \bbtitle{Proceedings of the IEEE Conference on Computer Vision and Pattern
  Recognition},
pp. \bfpage{7014}--\blpage{7023}
(\byear{2018})
\end{bchapter}
\endbibitem

\bibitem{zhou2016learning}
\begin{bchapter}
\bauthor{\bsnm{Zhou}, \binits{B.}},
\bauthor{\bsnm{Khosla}, \binits{A.}},
\bauthor{\bsnm{Lapedriza}, \binits{A.}},
\bauthor{\bsnm{Oliva}, \binits{A.}},
\bauthor{\bsnm{Torralba}, \binits{A.}}:
\bctitle{Learning deep features for discriminative localization}.
In: \bbtitle{Proceedings of the IEEE Conference on Computer Vision and Pattern
  Recognition},
pp. \bfpage{2921}--\blpage{2929}
(\byear{2016})
\end{bchapter}
\endbibitem

\bibitem{xu2022multi}
\begin{bchapter}
\bauthor{\bsnm{Xu}, \binits{L.}},
\bauthor{\bsnm{Ouyang}, \binits{W.}},
\bauthor{\bsnm{Bennamoun}, \binits{M.}},
\bauthor{\bsnm{Boussaid}, \binits{F.}},
\bauthor{\bsnm{Xu}, \binits{D.}}:
\bctitle{Multi-class token transformer for weakly supervised semantic
  segmentation}.
In: \bbtitle{Proceedings of the IEEE/CVF Conference on Computer Vision and
  Pattern Recognition},
pp. \bfpage{4310}--\blpage{4319}
(\byear{2022})
\end{bchapter}
\endbibitem

\bibitem{dosovitskiy2020image}
\begin{botherref}
\oauthor{\bsnm{Dosovitskiy}, \binits{A.}},
\oauthor{\bsnm{Beyer}, \binits{L.}},
\oauthor{\bsnm{Kolesnikov}, \binits{A.}},
\oauthor{\bsnm{Weissenborn}, \binits{D.}},
\oauthor{\bsnm{Zhai}, \binits{X.}},
\oauthor{\bsnm{Unterthiner}, \binits{T.}},
\oauthor{\bsnm{Dehghani}, \binits{M.}},
\oauthor{\bsnm{Minderer}, \binits{M.}},
\oauthor{\bsnm{Heigold}, \binits{G.}},
\oauthor{\bsnm{Gelly}, \binits{S.}}, et al.:
An image is worth 16x16 words: Transformers for image recognition at scale.
arXiv preprint arXiv:2010.11929
(2020)
\end{botherref}
\endbibitem

\bibitem{xie2021segformer}
\begin{botherref}
\oauthor{\bsnm{Xie}, \binits{E.}},
\oauthor{\bsnm{Wang}, \binits{W.}},
\oauthor{\bsnm{Yu}, \binits{Z.}},
\oauthor{\bsnm{Anandkumar}, \binits{A.}},
\oauthor{\bsnm{Alvarez}, \binits{J.M.}},
\oauthor{\bsnm{Luo}, \binits{P.}}:
Segformer: Simple and efficient design for semantic segmentation with
  transformers.
Advances in Neural Information Processing Systems
\textbf{34}
(2021)
\end{botherref}
\endbibitem

\bibitem{wang2021pyramid}
\begin{bchapter}
\bauthor{\bsnm{Wang}, \binits{W.}},
\bauthor{\bsnm{Xie}, \binits{E.}},
\bauthor{\bsnm{Li}, \binits{X.}},
\bauthor{\bsnm{Fan}, \binits{D.-P.}},
\bauthor{\bsnm{Song}, \binits{K.}},
\bauthor{\bsnm{Liang}, \binits{D.}},
\bauthor{\bsnm{Lu}, \binits{T.}},
\bauthor{\bsnm{Luo}, \binits{P.}},
\bauthor{\bsnm{Shao}, \binits{L.}}:
\bctitle{Pyramid vision transformer: A versatile backbone for dense prediction
  without convolutions}.
In: \bbtitle{Proceedings of the IEEE/CVF International Conference on Computer
  Vision},
pp. \bfpage{568}--\blpage{578}
(\byear{2021})
\end{bchapter}
\endbibitem

\bibitem{peng2021conformer}
\begin{bchapter}
\bauthor{\bsnm{Peng}, \binits{Z.}},
\bauthor{\bsnm{Huang}, \binits{W.}},
\bauthor{\bsnm{Gu}, \binits{S.}},
\bauthor{\bsnm{Xie}, \binits{L.}},
\bauthor{\bsnm{Wang}, \binits{Y.}},
\bauthor{\bsnm{Jiao}, \binits{J.}},
\bauthor{\bsnm{Ye}, \binits{Q.}}:
\bctitle{Conformer: Local features coupling global representations for visual
  recognition}.
In: \bbtitle{Proceedings of the IEEE/CVF International Conference on Computer
  Vision},
pp. \bfpage{367}--\blpage{376}
(\byear{2021})
\end{bchapter}
\endbibitem

\bibitem{li2022uniformer}
\begin{botherref}
\oauthor{\bsnm{Li}, \binits{K.}},
\oauthor{\bsnm{Wang}, \binits{Y.}},
\oauthor{\bsnm{Zhang}, \binits{J.}},
\oauthor{\bsnm{Gao}, \binits{P.}},
\oauthor{\bsnm{Song}, \binits{G.}},
\oauthor{\bsnm{Liu}, \binits{Y.}},
\oauthor{\bsnm{Li}, \binits{H.}},
\oauthor{\bsnm{Qiao}, \binits{Y.}}:
Uniformer: Unifying convolution and self-attention for visual recognition.
arXiv preprint arXiv:2201.09450
(2022)
\end{botherref}
\endbibitem

\bibitem{li2022transcam}
\begin{botherref}
\oauthor{\bsnm{Li}, \binits{R.}},
\oauthor{\bsnm{Mai}, \binits{Z.}},
\oauthor{\bsnm{Trabelsi}, \binits{C.}},
\oauthor{\bsnm{Zhang}, \binits{Z.}},
\oauthor{\bsnm{Jang}, \binits{J.}},
\oauthor{\bsnm{Sanner}, \binits{S.}}:
Transcam: Transformer attention-based cam refinement for weakly supervised
  semantic segmentation.
arXiv preprint arXiv:2203.07239
(2022)
\end{botherref}
\endbibitem

\bibitem{everingham2015pascal}
\begin{barticle}
\bauthor{\bsnm{Everingham}, \binits{M.}},
\bauthor{\bsnm{Eslami}, \binits{S.}},
\bauthor{\bsnm{Van~Gool}, \binits{L.}},
\bauthor{\bsnm{Williams}, \binits{C.K.}},
\bauthor{\bsnm{Winn}, \binits{J.}},
\bauthor{\bsnm{Zisserman}, \binits{A.}}:
\batitle{The pascal visual object classes challenge: A retrospective}.
\bjtitle{International journal of computer vision}
\bvolume{111}(\bissue{1}),
\bfpage{98}--\blpage{136}
(\byear{2015})
\end{barticle}
\endbibitem

\bibitem{simonyan2014very}
\begin{botherref}
\oauthor{\bsnm{Simonyan}, \binits{K.}},
\oauthor{\bsnm{Zisserman}, \binits{A.}}:
Very deep convolutional networks for large-scale image recognition.
arXiv preprint arXiv:1409.1556
(2014)
\end{botherref}
\endbibitem

\bibitem{he2016deep}
\begin{bchapter}
\bauthor{\bsnm{He}, \binits{K.}},
\bauthor{\bsnm{Zhang}, \binits{X.}},
\bauthor{\bsnm{Ren}, \binits{S.}},
\bauthor{\bsnm{Sun}, \binits{J.}}:
\bctitle{Deep residual learning for image recognition}.
In: \bbtitle{Proceedings of the IEEE Conference on Computer Vision and Pattern
  Recognition},
pp. \bfpage{770}--\blpage{778}
(\byear{2016})
\end{bchapter}
\endbibitem

\bibitem{hou2018self}
\begin{botherref}
\oauthor{\bsnm{Hou}, \binits{Q.}},
\oauthor{\bsnm{Jiang}, \binits{P.}},
\oauthor{\bsnm{Wei}, \binits{Y.}},
\oauthor{\bsnm{Cheng}, \binits{M.-M.}}:
Self-erasing network for integral object attention.
Advances in Neural Information Processing Systems
\textbf{31}
(2018)
\end{botherref}
\endbibitem

\bibitem{wei2018revisiting}
\begin{bchapter}
\bauthor{\bsnm{Wei}, \binits{Y.}},
\bauthor{\bsnm{Xiao}, \binits{H.}},
\bauthor{\bsnm{Shi}, \binits{H.}},
\bauthor{\bsnm{Jie}, \binits{Z.}},
\bauthor{\bsnm{Feng}, \binits{J.}},
\bauthor{\bsnm{Huang}, \binits{T.S.}}:
\bctitle{Revisiting dilated convolution: A simple approach for weakly-and
  semi-supervised semantic segmentation}.
In: \bbtitle{Proceedings of the IEEE Conference on Computer Vision and Pattern
  Recognition},
pp. \bfpage{7268}--\blpage{7277}
(\byear{2018})
\end{bchapter}
\endbibitem

\bibitem{lee2019ficklenet}
\begin{bchapter}
\bauthor{\bsnm{Lee}, \binits{J.}},
\bauthor{\bsnm{Kim}, \binits{E.}},
\bauthor{\bsnm{Lee}, \binits{S.}},
\bauthor{\bsnm{Lee}, \binits{J.}},
\bauthor{\bsnm{Yoon}, \binits{S.}}:
\bctitle{Ficklenet: Weakly and semi-supervised semantic image segmentation
  using stochastic inference}.
In: \bbtitle{Proceedings of the IEEE/CVF Conference on Computer Vision and
  Pattern Recognition},
pp. \bfpage{5267}--\blpage{5276}
(\byear{2019})
\end{bchapter}
\endbibitem

\bibitem{jiang2019integral}
\begin{bchapter}
\bauthor{\bsnm{Jiang}, \binits{P.-T.}},
\bauthor{\bsnm{Hou}, \binits{Q.}},
\bauthor{\bsnm{Cao}, \binits{Y.}},
\bauthor{\bsnm{Cheng}, \binits{M.-M.}},
\bauthor{\bsnm{Wei}, \binits{Y.}},
\bauthor{\bsnm{Xiong}, \binits{H.-K.}}:
\bctitle{Integral object mining via online attention accumulation}.
In: \bbtitle{Proceedings of the IEEE/CVF International Conference on Computer
  Vision},
pp. \bfpage{2070}--\blpage{2079}
(\byear{2019})
\end{bchapter}
\endbibitem

\bibitem{chang2020weakly}
\begin{bchapter}
\bauthor{\bsnm{Chang}, \binits{Y.-T.}},
\bauthor{\bsnm{Wang}, \binits{Q.}},
\bauthor{\bsnm{Hung}, \binits{W.-C.}},
\bauthor{\bsnm{Piramuthu}, \binits{R.}},
\bauthor{\bsnm{Tsai}, \binits{Y.-H.}},
\bauthor{\bsnm{Yang}, \binits{M.-H.}}:
\bctitle{Weakly-supervised semantic segmentation via sub-category exploration}.
In: \bbtitle{Proceedings of the IEEE/CVF Conference on Computer Vision and
  Pattern Recognition},
pp. \bfpage{8991}--\blpage{9000}
(\byear{2020})
\end{bchapter}
\endbibitem

\bibitem{chen2022self}
\begin{bchapter}
\bauthor{\bsnm{Chen}, \binits{Q.}},
\bauthor{\bsnm{Yang}, \binits{L.}},
\bauthor{\bsnm{Lai}, \binits{J.-H.}},
\bauthor{\bsnm{Xie}, \binits{X.}}:
\bctitle{Self-supervised image-specific prototype exploration for weakly
  supervised semantic segmentation}.
In: \bbtitle{Proceedings of the IEEE/CVF Conference on Computer Vision and
  Pattern Recognition},
pp. \bfpage{4288}--\blpage{4298}
(\byear{2022})
\end{bchapter}
\endbibitem

\bibitem{sun2020mining}
\begin{bchapter}
\bauthor{\bsnm{Sun}, \binits{G.}},
\bauthor{\bsnm{Wang}, \binits{W.}},
\bauthor{\bsnm{Dai}, \binits{J.}},
\bauthor{\bsnm{Van~Gool}, \binits{L.}}:
\bctitle{Mining cross-image semantics for weakly supervised semantic
  segmentation}.
In: \bbtitle{European Conference on Computer Vision},
pp. \bfpage{347}--\blpage{365}
(\byear{2020}).
\bcomment{Springer}
\end{bchapter}
\endbibitem

\bibitem{kumar2017hide}
\begin{bchapter}
\bauthor{\bsnm{Kumar~Singh}, \binits{K.}},
\bauthor{\bsnm{Jae~Lee}, \binits{Y.}}:
\bctitle{Hide-and-seek: Forcing a network to be meticulous for
  weakly-supervised object and action localization}.
In: \bbtitle{Proceedings of the IEEE International Conference on Computer
  Vision},
pp. \bfpage{3524}--\blpage{3533}
(\byear{2017})
\end{bchapter}
\endbibitem

\bibitem{zhang2021complementary}
\begin{bchapter}
\bauthor{\bsnm{Zhang}, \binits{F.}},
\bauthor{\bsnm{Gu}, \binits{C.}},
\bauthor{\bsnm{Zhang}, \binits{C.}},
\bauthor{\bsnm{Dai}, \binits{Y.}}:
\bctitle{Complementary patch for weakly supervised semantic segmentation}.
In: \bbtitle{Proceedings of the IEEE/CVF International Conference on Computer
  Vision},
pp. \bfpage{7242}--\blpage{7251}
(\byear{2021})
\end{bchapter}
\endbibitem

\bibitem{ranftl2021vision}
\begin{bchapter}
\bauthor{\bsnm{Ranftl}, \binits{R.}},
\bauthor{\bsnm{Bochkovskiy}, \binits{A.}},
\bauthor{\bsnm{Koltun}, \binits{V.}}:
\bctitle{Vision transformers for dense prediction}.
In: \bbtitle{Proceedings of the IEEE/CVF International Conference on Computer
  Vision},
pp. \bfpage{12179}--\blpage{12188}
(\byear{2021})
\end{bchapter}
\endbibitem

\bibitem{arnab2021vivit}
\begin{bchapter}
\bauthor{\bsnm{Arnab}, \binits{A.}},
\bauthor{\bsnm{Dehghani}, \binits{M.}},
\bauthor{\bsnm{Heigold}, \binits{G.}},
\bauthor{\bsnm{Sun}, \binits{C.}},
\bauthor{\bsnm{Lu{\v{c}}i{\'c}}, \binits{M.}},
\bauthor{\bsnm{Schmid}, \binits{C.}}:
\bctitle{Vivit: A video vision transformer}.
In: \bbtitle{Proceedings of the IEEE/CVF International Conference on Computer
  Vision},
pp. \bfpage{6836}--\blpage{6846}
(\byear{2021})
\end{bchapter}
\endbibitem

\bibitem{ru2022learning}
\begin{bchapter}
\bauthor{\bsnm{Ru}, \binits{L.}},
\bauthor{\bsnm{Zhan}, \binits{Y.}},
\bauthor{\bsnm{Yu}, \binits{B.}},
\bauthor{\bsnm{Du}, \binits{B.}}:
\bctitle{Learning affinity from attention: End-to-end weakly-supervised
  semantic segmentation with transformers}.
In: \bbtitle{Proceedings of the IEEE/CVF Conference on Computer Vision and
  Pattern Recognition},
pp. \bfpage{16846}--\blpage{16855}
(\byear{2022})
\end{bchapter}
\endbibitem

\bibitem{ke2021universal}
\begin{botherref}
\oauthor{\bsnm{Ke}, \binits{T.-W.}},
\oauthor{\bsnm{Hwang}, \binits{J.-J.}},
\oauthor{\bsnm{Yu}, \binits{S.X.}}:
Universal weakly supervised segmentation by pixel-to-segment contrastive
  learning.
arXiv preprint arXiv:2105.00957
(2021)
\end{botherref}
\endbibitem

\bibitem{kolesnikov2016seed}
\begin{bchapter}
\bauthor{\bsnm{Kolesnikov}, \binits{A.}},
\bauthor{\bsnm{Lampert}, \binits{C.H.}}:
\bctitle{Seed, expand and constrain: Three principles for weakly-supervised
  image segmentation}.
In: \bbtitle{European Conference on Computer Vision},
pp. \bfpage{695}--\blpage{711}
(\byear{2016}).
\bcomment{Springer}
\end{bchapter}
\endbibitem

\bibitem{zhang2020reliability}
\begin{bchapter}
\bauthor{\bsnm{Zhang}, \binits{B.}},
\bauthor{\bsnm{Xiao}, \binits{J.}},
\bauthor{\bsnm{Wei}, \binits{Y.}},
\bauthor{\bsnm{Sun}, \binits{M.}},
\bauthor{\bsnm{Huang}, \binits{K.}}:
\bctitle{Reliability does matter: An end-to-end weakly supervised semantic
  segmentation approach}.
In: \bbtitle{Proceedings of the AAAI Conference on Artificial Intelligence},
vol. \bseriesno{34},
pp. \bfpage{12765}--\blpage{12772}
(\byear{2020})
\end{bchapter}
\endbibitem

\bibitem{kim2021discriminative}
\begin{bchapter}
\bauthor{\bsnm{Kim}, \binits{B.}},
\bauthor{\bsnm{Han}, \binits{S.}},
\bauthor{\bsnm{Kim}, \binits{J.}}:
\bctitle{Discriminative region suppression for weakly-supervised semantic
  segmentation}.
In: \bbtitle{Proceedings of the AAAI Conference on Artificial Intelligence},
vol. \bseriesno{35},
pp. \bfpage{1754}--\blpage{1761}
(\byear{2021})
\end{bchapter}
\endbibitem

\bibitem{yao2021non}
\begin{bchapter}
\bauthor{\bsnm{Yao}, \binits{Y.}},
\bauthor{\bsnm{Chen}, \binits{T.}},
\bauthor{\bsnm{Xie}, \binits{G.-S.}},
\bauthor{\bsnm{Zhang}, \binits{C.}},
\bauthor{\bsnm{Shen}, \binits{F.}},
\bauthor{\bsnm{Wu}, \binits{Q.}},
\bauthor{\bsnm{Tang}, \binits{Z.}},
\bauthor{\bsnm{Zhang}, \binits{J.}}:
\bctitle{Non-salient region object mining for weakly supervised semantic
  segmentation}.
In: \bbtitle{Proceedings of the IEEE/CVF Conference on Computer Vision and
  Pattern Recognition},
pp. \bfpage{2623}--\blpage{2632}
(\byear{2021})
\end{bchapter}
\endbibitem

\bibitem{wang2019self}
\begin{botherref}
\oauthor{\bsnm{Wang}, \binits{Y.}},
\oauthor{\bsnm{Zhang}, \binits{J.}},
\oauthor{\bsnm{Kan}, \binits{M.}},
\oauthor{\bsnm{Shan}, \binits{S.}},
\oauthor{\bsnm{Chen}, \binits{X.}}:
Self-supervised scale equivariant network for weakly supervised semantic
  segmentation.
arXiv preprint arXiv:1909.03714
(2019)
\end{botherref}
\endbibitem

\bibitem{wang2020self}
\begin{bchapter}
\bauthor{\bsnm{Wang}, \binits{Y.}},
\bauthor{\bsnm{Zhang}, \binits{J.}},
\bauthor{\bsnm{Kan}, \binits{M.}},
\bauthor{\bsnm{Shan}, \binits{S.}},
\bauthor{\bsnm{Chen}, \binits{X.}}:
\bctitle{Self-supervised equivariant attention mechanism for weakly supervised
  semantic segmentation}.
In: \bbtitle{Proceedings of the IEEE/CVF Conference on Computer Vision and
  Pattern Recognition},
pp. \bfpage{12275}--\blpage{12284}
(\byear{2020})
\end{bchapter}
\endbibitem

\bibitem{fan2020cian}
\begin{bchapter}
\bauthor{\bsnm{Fan}, \binits{J.}},
\bauthor{\bsnm{Zhang}, \binits{Z.}},
\bauthor{\bsnm{Tan}, \binits{T.}},
\bauthor{\bsnm{Song}, \binits{C.}},
\bauthor{\bsnm{Xiao}, \binits{J.}}:
\bctitle{Cian: Cross-image affinity net for weakly supervised semantic
  segmentation}.
In: \bbtitle{Proceedings of the AAAI Conference on Artificial Intelligence},
vol. \bseriesno{34},
pp. \bfpage{10762}--\blpage{10769}
(\byear{2020})
\end{bchapter}
\endbibitem

\bibitem{chen2022class}
\begin{bchapter}
\bauthor{\bsnm{Chen}, \binits{Z.}},
\bauthor{\bsnm{Wang}, \binits{T.}},
\bauthor{\bsnm{Wu}, \binits{X.}},
\bauthor{\bsnm{Hua}, \binits{X.-S.}},
\bauthor{\bsnm{Zhang}, \binits{H.}},
\bauthor{\bsnm{Sun}, \binits{Q.}}:
\bctitle{Class re-activation maps for weakly-supervised semantic segmentation}.
In: \bbtitle{Proceedings of the IEEE/CVF Conference on Computer Vision and
  Pattern Recognition},
pp. \bfpage{969}--\blpage{978}
(\byear{2022})
\end{bchapter}
\endbibitem

\bibitem{li2022expansion}
\begin{botherref}
\oauthor{\bsnm{Li}, \binits{J.}},
\oauthor{\bsnm{Jie}, \binits{Z.}},
\oauthor{\bsnm{Wang}, \binits{X.}},
\oauthor{\bsnm{Wei}, \binits{X.}},
\oauthor{\bsnm{Ma}, \binits{L.}}:
Expansion and shrinkage of localization for weakly-supervised semantic
  segmentation.
arXiv preprint arXiv:2209.07761
(2022)
\end{botherref}
\endbibitem

\bibitem{roy2017combining}
\begin{bchapter}
\bauthor{\bsnm{Roy}, \binits{A.}},
\bauthor{\bsnm{Todorovic}, \binits{S.}}:
\bctitle{Combining bottom-up, top-down, and smoothness cues for weakly
  supervised image segmentation}.
In: \bbtitle{Proceedings of the IEEE Conference on Computer Vision and Pattern
  Recognition},
pp. \bfpage{3529}--\blpage{3538}
(\byear{2017})
\end{bchapter}
\endbibitem

\bibitem{chaudhry2017discovering}
\begin{botherref}
\oauthor{\bsnm{Chaudhry}, \binits{A.}},
\oauthor{\bsnm{Dokania}, \binits{P.K.}},
\oauthor{\bsnm{Torr}, \binits{P.H.}}:
Discovering class-specific pixels for weakly-supervised semantic segmentation.
arXiv preprint arXiv:1707.05821
(2017)
\end{botherref}
\endbibitem

\bibitem{sun2022inferring}
\begin{bchapter}
\bauthor{\bsnm{Sun}, \binits{W.}},
\bauthor{\bsnm{Zhang}, \binits{J.}},
\bauthor{\bsnm{Barnes}, \binits{N.}}:
\bctitle{Inferring the class conditional response map for weakly supervised
  semantic segmentation}.
In: \bbtitle{Proceedings of the IEEE/CVF Winter Conference on Applications of
  Computer Vision},
pp. \bfpage{2878}--\blpage{2887}
(\byear{2022})
\end{bchapter}
\endbibitem

\bibitem{li2022uncertainty}
\begin{bchapter}
\bauthor{\bsnm{Li}, \binits{Y.}},
\bauthor{\bsnm{Duan}, \binits{Y.}},
\bauthor{\bsnm{Kuang}, \binits{Z.}},
\bauthor{\bsnm{Chen}, \binits{Y.}},
\bauthor{\bsnm{Zhang}, \binits{W.}},
\bauthor{\bsnm{Li}, \binits{X.}}:
\bctitle{Uncertainty estimation via response scaling for pseudo-mask noise
  mitigation in weakly-supervised semantic segmentation}.
In: \bbtitle{Proceedings of the AAAI Conference on Artificial Intelligence},
vol. \bseriesno{36},
pp. \bfpage{1447}--\blpage{1455}
(\byear{2022})
\end{bchapter}
\endbibitem

\bibitem{yu2022ex}
\begin{botherref}
\oauthor{\bsnm{Yu}, \binits{L.}},
\oauthor{\bsnm{Xiang}, \binits{W.}},
\oauthor{\bsnm{Fang}, \binits{J.}},
\oauthor{\bsnm{Chen}, \binits{Y.-P.P.}},
\oauthor{\bsnm{Chi}, \binits{L.}}:
ex-vit: A novel explainable vision transformer for weakly supervised semantic
  segmentation.
arXiv preprint arXiv:2207.05358
(2022)
\end{botherref}
\endbibitem

\bibitem{huang2022attention}
\begin{botherref}
\oauthor{\bsnm{Huang}, \binits{J.}},
\oauthor{\bsnm{Wang}, \binits{J.}},
\oauthor{\bsnm{Sun}, \binits{Q.}},
\oauthor{\bsnm{Zhang}, \binits{H.}}:
Attention-based class activation diffusion for weakly-supervised semantic
  segmentation.
arXiv preprint arXiv:2211.10931
(2022)
\end{botherref}
\endbibitem

\bibitem{chen2022semformer}
\begin{botherref}
\oauthor{\bsnm{Chen}, \binits{J.}},
\oauthor{\bsnm{Zhao}, \binits{X.}},
\oauthor{\bsnm{Luo}, \binits{C.}},
\oauthor{\bsnm{Shen}, \binits{L.}}:
Semformer: Semantic guided activation transformer for weakly supervised
  semantic segmentation.
arXiv preprint arXiv:2210.14618
(2022)
\end{botherref}
\endbibitem

\bibitem{achanta2012slic}
\begin{barticle}
\bauthor{\bsnm{Achanta}, \binits{R.}},
\bauthor{\bsnm{Shaji}, \binits{A.}},
\bauthor{\bsnm{Smith}, \binits{K.}},
\bauthor{\bsnm{Lucchi}, \binits{A.}},
\bauthor{\bsnm{Fua}, \binits{P.}},
\bauthor{\bsnm{S{\"u}sstrunk}, \binits{S.}}:
\batitle{Slic superpixels compared to state-of-the-art superpixel methods}.
\bjtitle{IEEE transactions on pattern analysis and machine intelligence}
\bvolume{34}(\bissue{11}),
\bfpage{2274}--\blpage{2282}
(\byear{2012})
\end{barticle}
\endbibitem

\bibitem{lin2014microsoft}
\begin{bchapter}
\bauthor{\bsnm{Lin}, \binits{T.-Y.}},
\bauthor{\bsnm{Maire}, \binits{M.}},
\bauthor{\bsnm{Belongie}, \binits{S.}},
\bauthor{\bsnm{Hays}, \binits{J.}},
\bauthor{\bsnm{Perona}, \binits{P.}},
\bauthor{\bsnm{Ramanan}, \binits{D.}},
\bauthor{\bsnm{Doll{\'a}r}, \binits{P.}},
\bauthor{\bsnm{Zitnick}, \binits{C.L.}}:
\bctitle{Microsoft coco: Common objects in context}.
In: \bbtitle{Computer Vision--ECCV 2014: 13th European Conference, Zurich,
  Switzerland, September 6-12, 2014, Proceedings, Part V 13},
pp. \bfpage{740}--\blpage{755}
(\byear{2014}).
\bcomment{Springer}
\end{bchapter}
\endbibitem

\bibitem{hariharan2011semantic}
\begin{bchapter}
\bauthor{\bsnm{Hariharan}, \binits{B.}},
\bauthor{\bsnm{Arbel{\'a}ez}, \binits{P.}},
\bauthor{\bsnm{Bourdev}, \binits{L.}},
\bauthor{\bsnm{Maji}, \binits{S.}},
\bauthor{\bsnm{Malik}, \binits{J.}}:
\bctitle{Semantic contours from inverse detectors}.
In: \bbtitle{2011 International Conference on Computer Vision},
pp. \bfpage{991}--\blpage{998}
(\byear{2011}).
\bcomment{IEEE}
\end{bchapter}
\endbibitem

\bibitem{russakovsky2015imagenet}
\begin{barticle}
\bauthor{\bsnm{Russakovsky}, \binits{O.}},
\bauthor{\bsnm{Deng}, \binits{J.}},
\bauthor{\bsnm{Su}, \binits{H.}},
\bauthor{\bsnm{Krause}, \binits{J.}},
\bauthor{\bsnm{Satheesh}, \binits{S.}},
\bauthor{\bsnm{Ma}, \binits{S.}},
\bauthor{\bsnm{Huang}, \binits{Z.}},
\bauthor{\bsnm{Karpathy}, \binits{A.}},
\bauthor{\bsnm{Khosla}, \binits{A.}},
\bauthor{\bsnm{Bernstein}, \binits{M.}}, \betal:
\batitle{Imagenet large scale visual recognition challenge}.
\bjtitle{International journal of computer vision}
\bvolume{115}(\bissue{3}),
\bfpage{211}--\blpage{252}
(\byear{2015})
\end{barticle}
\endbibitem

\bibitem{chen2017deeplab}
\begin{barticle}
\bauthor{\bsnm{Chen}, \binits{L.-C.}},
\bauthor{\bsnm{Papandreou}, \binits{G.}},
\bauthor{\bsnm{Kokkinos}, \binits{I.}},
\bauthor{\bsnm{Murphy}, \binits{K.}},
\bauthor{\bsnm{Yuille}, \binits{A.L.}}:
\batitle{Deeplab: Semantic image segmentation with deep convolutional nets,
  atrous convolution, and fully connected crfs}.
\bjtitle{IEEE transactions on pattern analysis and machine intelligence}
\bvolume{40}(\bissue{4}),
\bfpage{834}--\blpage{848}
(\byear{2017})
\end{barticle}
\endbibitem

\bibitem{wu2019wider}
\begin{barticle}
\bauthor{\bsnm{Wu}, \binits{Z.}},
\bauthor{\bsnm{Shen}, \binits{C.}},
\bauthor{\bsnm{Van Den~Hengel}, \binits{A.}}:
\batitle{Wider or deeper: Revisiting the resnet model for visual recognition}.
\bjtitle{Pattern Recognition}
\bvolume{90},
\bfpage{119}--\blpage{133}
(\byear{2019})
\end{barticle}
\endbibitem

\bibitem{lee2021anti}
\begin{bchapter}
\bauthor{\bsnm{Lee}, \binits{J.}},
\bauthor{\bsnm{Kim}, \binits{E.}},
\bauthor{\bsnm{Yoon}, \binits{S.}}:
\bctitle{Anti-adversarially manipulated attributions for weakly and
  semi-supervised semantic segmentation}.
In: \bbtitle{Proceedings of the IEEE/CVF Conference on Computer Vision and
  Pattern Recognition},
pp. \bfpage{4071}--\blpage{4080}
(\byear{2021})
\end{bchapter}
\endbibitem

\bibitem{qin2022activation}
\begin{bchapter}
\bauthor{\bsnm{Qin}, \binits{J.}},
\bauthor{\bsnm{Wu}, \binits{J.}},
\bauthor{\bsnm{Xiao}, \binits{X.}},
\bauthor{\bsnm{Li}, \binits{L.}},
\bauthor{\bsnm{Wang}, \binits{X.}}:
\bctitle{Activation modulation and recalibration scheme for weakly supervised
  semantic segmentation}.
In: \bbtitle{Proceedings of the AAAI Conference on Artificial Intelligence},
vol. \bseriesno{36},
pp. \bfpage{2117}--\blpage{2125}
(\byear{2022})
\end{bchapter}
\endbibitem

\bibitem{chen2022saliency}
\begin{botherref}
\oauthor{\bsnm{Chen}, \binits{T.}},
\oauthor{\bsnm{Yao}, \binits{Y.}},
\oauthor{\bsnm{Zhang}, \binits{L.}},
\oauthor{\bsnm{Wang}, \binits{Q.}},
\oauthor{\bsnm{Xie}, \binits{G.}},
\oauthor{\bsnm{Shen}, \binits{F.}}:
Saliency guided inter-and intra-class relation constraints for weakly
  supervised semantic segmentation.
IEEE Transactions on Multimedia
(2022)
\end{botherref}
\endbibitem

\bibitem{chen2020weakly}
\begin{bchapter}
\bauthor{\bsnm{Chen}, \binits{L.}},
\bauthor{\bsnm{Wu}, \binits{W.}},
\bauthor{\bsnm{Fu}, \binits{C.}},
\bauthor{\bsnm{Han}, \binits{X.}},
\bauthor{\bsnm{Zhang}, \binits{Y.}}:
\bctitle{Weakly supervised semantic segmentation with boundary exploration}.
In: \bbtitle{European Conference on Computer Vision},
pp. \bfpage{347}--\blpage{362}
(\byear{2020}).
\bcomment{Springer}
\end{bchapter}
\endbibitem

\bibitem{sun2021ecs}
\begin{bchapter}
\bauthor{\bsnm{Sun}, \binits{K.}},
\bauthor{\bsnm{Shi}, \binits{H.}},
\bauthor{\bsnm{Zhang}, \binits{Z.}},
\bauthor{\bsnm{Huang}, \binits{Y.}}:
\bctitle{Ecs-net: Improving weakly supervised semantic segmentation by using
  connections between class activation maps}.
In: \bbtitle{Proceedings of the IEEE/CVF International Conference on Computer
  Vision},
pp. \bfpage{7283}--\blpage{7292}
(\byear{2021})
\end{bchapter}
\endbibitem

\bibitem{li2018tell}
\begin{bchapter}
\bauthor{\bsnm{Li}, \binits{K.}},
\bauthor{\bsnm{Wu}, \binits{Z.}},
\bauthor{\bsnm{Peng}, \binits{K.-C.}},
\bauthor{\bsnm{Ernst}, \binits{J.}},
\bauthor{\bsnm{Fu}, \binits{Y.}}:
\bctitle{Tell me where to look: Guided attention inference network}.
In: \bbtitle{Proceedings of the IEEE Conference on Computer Vision and Pattern
  Recognition},
pp. \bfpage{9215}--\blpage{9223}
(\byear{2018})
\end{bchapter}
\endbibitem

\bibitem{wang2018weakly}
\begin{bchapter}
\bauthor{\bsnm{Wang}, \binits{X.}},
\bauthor{\bsnm{You}, \binits{S.}},
\bauthor{\bsnm{Li}, \binits{X.}},
\bauthor{\bsnm{Ma}, \binits{H.}}:
\bctitle{Weakly-supervised semantic segmentation by iteratively mining common
  object features}.
In: \bbtitle{Proceedings of the IEEE Conference on Computer Vision and Pattern
  Recognition},
pp. \bfpage{1354}--\blpage{1362}
(\byear{2018})
\end{bchapter}
\endbibitem

\bibitem{fan2020learning}
\begin{bchapter}
\bauthor{\bsnm{Fan}, \binits{J.}},
\bauthor{\bsnm{Zhang}, \binits{Z.}},
\bauthor{\bsnm{Song}, \binits{C.}},
\bauthor{\bsnm{Tan}, \binits{T.}}:
\bctitle{Learning integral objects with intra-class discriminator for
  weakly-supervised semantic segmentation}.
In: \bbtitle{Proceedings of the IEEE/CVF Conference on Computer Vision and
  Pattern Recognition},
pp. \bfpage{4283}--\blpage{4292}
(\byear{2020})
\end{bchapter}
\endbibitem

\bibitem{yao2020saliency}
\begin{barticle}
\bauthor{\bsnm{Yao}, \binits{Q.}},
\bauthor{\bsnm{Gong}, \binits{X.}}:
\batitle{Saliency guided self-attention network for weakly and semi-supervised
  semantic segmentation}.
\bjtitle{IEEE Access}
\bvolume{8},
\bfpage{14413}--\blpage{14423}
(\byear{2020})
\end{barticle}
\endbibitem

\bibitem{zhang2020causal}
\begin{barticle}
\bauthor{\bsnm{Zhang}, \binits{D.}},
\bauthor{\bsnm{Zhang}, \binits{H.}},
\bauthor{\bsnm{Tang}, \binits{J.}},
\bauthor{\bsnm{Hua}, \binits{X.-S.}},
\bauthor{\bsnm{Sun}, \binits{Q.}}:
\batitle{Causal intervention for weakly-supervised semantic segmentation}.
\bjtitle{Advances in Neural Information Processing Systems}
\bvolume{33},
\bfpage{655}--\blpage{666}
(\byear{2020})
\end{barticle}
\endbibitem

\bibitem{kweon2021unlocking}
\begin{bchapter}
\bauthor{\bsnm{Kweon}, \binits{H.}},
\bauthor{\bsnm{Yoon}, \binits{S.-H.}},
\bauthor{\bsnm{Kim}, \binits{H.}},
\bauthor{\bsnm{Park}, \binits{D.}},
\bauthor{\bsnm{Yoon}, \binits{K.-J.}}:
\bctitle{Unlocking the potential of ordinary classifier: Class-specific
  adversarial erasing framework for weakly supervised semantic segmentation}.
In: \bbtitle{Proceedings of the IEEE/CVF International Conference on Computer
  Vision},
pp. \bfpage{6994}--\blpage{7003}
(\byear{2021})
\end{bchapter}
\endbibitem

\bibitem{lee2022threshold}
\begin{bchapter}
\bauthor{\bsnm{Lee}, \binits{M.}},
\bauthor{\bsnm{Kim}, \binits{D.}},
\bauthor{\bsnm{Shim}, \binits{H.}}:
\bctitle{Threshold matters in wsss: manipulating the activation for the robust
  and accurate segmentation model against thresholds}.
In: \bbtitle{Proceedings of the IEEE/CVF Conference on Computer Vision and
  Pattern Recognition},
pp. \bfpage{4330}--\blpage{4339}
(\byear{2022})
\end{bchapter}
\endbibitem

\bibitem{xu2021leveraging}
\begin{bchapter}
\bauthor{\bsnm{Xu}, \binits{L.}},
\bauthor{\bsnm{Ouyang}, \binits{W.}},
\bauthor{\bsnm{Bennamoun}, \binits{M.}},
\bauthor{\bsnm{Boussaid}, \binits{F.}},
\bauthor{\bsnm{Sohel}, \binits{F.}},
\bauthor{\bsnm{Xu}, \binits{D.}}:
\bctitle{Leveraging auxiliary tasks with affinity learning for weakly
  supervised semantic segmentation}.
In: \bbtitle{Proceedings of the IEEE/CVF International Conference on Computer
  Vision},
pp. \bfpage{6984}--\blpage{6993}
(\byear{2021})
\end{bchapter}
\endbibitem

\bibitem{lee2021reducing}
\begin{barticle}
\bauthor{\bsnm{Lee}, \binits{J.}},
\bauthor{\bsnm{Choi}, \binits{J.}},
\bauthor{\bsnm{Mok}, \binits{J.}},
\bauthor{\bsnm{Yoon}, \binits{S.}}:
\batitle{Reducing information bottleneck for weakly supervised semantic
  segmentation}.
\bjtitle{Advances in Neural Information Processing Systems}
\bvolume{34},
\bfpage{27408}--\blpage{27421}
(\byear{2021})
\end{barticle}
\endbibitem

\bibitem{zeng2019joint}
\begin{bchapter}
\bauthor{\bsnm{Zeng}, \binits{Y.}},
\bauthor{\bsnm{Zhuge}, \binits{Y.}},
\bauthor{\bsnm{Lu}, \binits{H.}},
\bauthor{\bsnm{Zhang}, \binits{L.}}:
\bctitle{Joint learning of saliency detection and weakly supervised semantic
  segmentation}.
In: \bbtitle{Proceedings of the IEEE/CVF International Conference on Computer
  Vision},
pp. \bfpage{7223}--\blpage{7233}
(\byear{2019})
\end{bchapter}
\endbibitem

\bibitem{krahenbuhl2011efficient}
\begin{botherref}
\oauthor{\bsnm{Kr{\"a}henb{\"u}hl}, \binits{P.}},
\oauthor{\bsnm{Koltun}, \binits{V.}}:
Efficient inference in fully connected crfs with gaussian edge potentials.
Advances in neural information processing systems
\textbf{24}
(2011)
\end{botherref}
\endbibitem

\bibitem{wang2020weakly}
\begin{barticle}
\bauthor{\bsnm{Wang}, \binits{X.}},
\bauthor{\bsnm{Liu}, \binits{S.}},
\bauthor{\bsnm{Ma}, \binits{H.}},
\bauthor{\bsnm{Yang}, \binits{M.-H.}}:
\batitle{Weakly-supervised semantic segmentation by iterative affinity
  learning}.
\bjtitle{International Journal of Computer Vision}
\bvolume{128},
\bfpage{1736}--\blpage{1749}
(\byear{2020})
\end{barticle}
\endbibitem

\bibitem{luo2020learning}
\begin{bchapter}
\bauthor{\bsnm{Luo}, \binits{W.}},
\bauthor{\bsnm{Yang}, \binits{M.}}:
\bctitle{Learning saliency-free model with generic features for
  weakly-supervised semantic segmentation}.
In: \bbtitle{Proceedings of the AAAI Conference on Artificial Intelligence},
vol. \bseriesno{34},
pp. \bfpage{11717}--\blpage{11724}
(\byear{2020})
\end{bchapter}
\endbibitem

\bibitem{su2021context}
\begin{bchapter}
\bauthor{\bsnm{Su}, \binits{Y.}},
\bauthor{\bsnm{Sun}, \binits{R.}},
\bauthor{\bsnm{Lin}, \binits{G.}},
\bauthor{\bsnm{Wu}, \binits{Q.}}:
\bctitle{Context decoupling augmentation for weakly supervised semantic
  segmentation}.
In: \bbtitle{Proceedings of the IEEE/CVF International Conference on Computer
  Vision},
pp. \bfpage{7004}--\blpage{7014}
(\byear{2021})
\end{bchapter}
\endbibitem

\bibitem{pan2022learning}
\begin{barticle}
\bauthor{\bsnm{Pan}, \binits{J.}},
\bauthor{\bsnm{Zhu}, \binits{P.}},
\bauthor{\bsnm{Zhang}, \binits{K.}},
\bauthor{\bsnm{Cao}, \binits{B.}},
\bauthor{\bsnm{Wang}, \binits{Y.}},
\bauthor{\bsnm{Zhang}, \binits{D.}},
\bauthor{\bsnm{Han}, \binits{J.}},
\bauthor{\bsnm{Hu}, \binits{Q.}}:
\batitle{Learning self-supervised low-rank network for single-stage weakly and
  semi-supervised semantic segmentation}.
\bjtitle{International Journal of Computer Vision}
\bvolume{130}(\bissue{5}),
\bfpage{1181}--\blpage{1195}
(\byear{2022})
\end{barticle}
\endbibitem

\bibitem{zhou2021group}
\begin{barticle}
\bauthor{\bsnm{Zhou}, \binits{T.}},
\bauthor{\bsnm{Li}, \binits{L.}},
\bauthor{\bsnm{Li}, \binits{X.}},
\bauthor{\bsnm{Feng}, \binits{C.-M.}},
\bauthor{\bsnm{Li}, \binits{J.}},
\bauthor{\bsnm{Shao}, \binits{L.}}:
\batitle{Group-wise learning for weakly supervised semantic segmentation}.
\bjtitle{IEEE Transactions on Image Processing}
\bvolume{31},
\bfpage{799}--\blpage{811}
(\byear{2021})
\end{barticle}
\endbibitem

\end{thebibliography}


\end{document}